\documentclass[10pt,twocolumn,letterpaper]{article}

\usepackage{cvpr}      % To produce the REVIEW version

\usepackage[accsupp]{axessibility}
\usepackage{times}
\usepackage{graphicx}
\usepackage{amsmath}
\usepackage{amssymb}
\usepackage[dvipsnames, table, xcdraw]{xcolor}
\graphicspath{{./figures/}}

\usepackage{amssymb}
\usepackage{booktabs}
\usepackage{bm}
\usepackage{overpic}
\usepackage{multirow}
\usepackage{siunitx}
\usepackage{enumitem}
\usepackage{colortbl}
\usepackage{color}

\newcommand{\myPara}[1]{\vspace{6pt}\noindent\textbf{#1}}
\newcommand{\sArt}{state-of-the-art~}

\definecolor{ForestGreen}{RGB}{34,139,34}
\newcommand{\fg}[1]{\textbf{\textcolor{ForestGreen}{#1}}}
\definecolor{Blue}{RGB}{130, 176, 210}
\definecolor{Orange}{RGB}{255, 190, 122}
\definecolor{Salmon}{RGB}{255,140,105}

\newcommand{\addFig}[1]{}
\newcommand{\addFigs}[1]{}
\def\MyMthd{CrossKD}
\def\MyMthdL{Cross-Head Knowledge Distillation}
\def\bMyMthd{\textbf{\MyMthd{}}}
\def\cc{\cellcolor[gray]{.95}}

\definecolor{cvprblue}{rgb}{0.21,0.49,0.74}
\usepackage[pagebackref=false,breaklinks,colorlinks,citecolor=cvprblue]{hyperref}

\def\paperID{152} % *** Enter the Paper ID here
\def\confName{CVPR}
\def\confYear{2024}

\title{\MyMthd{}: Cross-Head Knowledge Distillation for Object Detection}

\author{Jiabao Wang$^{1}$\thanks{Equal contribution.}, Yuming Chen$^{1*}$, Zhaohui Zheng$^1$, Xiang Li$^{2,1}$, Ming-Ming Cheng$^{2,1}$, Qibin Hou$^{2,1}$\thanks{Corresponding author.} \\ \\
    $^1$VCIP, College of Computer Science, Nankai University \\
    $^2$NKIARI, Shenzhen Futian \\
    {\tt\small https://github.com/jbwang1997/CrossKD}
}

\begin{document}

\maketitle

\begin{abstract}

% Knowledge Distillation (KD) has been validated as an effective model compression technique for learning compact object detectors.
% %
% Existing state-of-the-art KD methods for object detection are mostly based on feature imitation, which is generally observed to be better than prediction mimicking.
% %
% In this paper, we observe that the inconsistency of the optimization objectives between the ground-truth signals and distillation targets may impair the performance of prediction mimicking.
% %
% To alleviate this issue, we present a general and effective distillation scheme, called \MyMthd{}, which delivers the intermediate features of the student's detection head to the teacher's detection head. 
% %
% The resulting cross-head predictions are then forced to mimic the teacher's predictions.
% %
% This manner relieves the student's head from receiving contradictory supervision signals from the annotations and the teacher's predictions, greatly improving the student's detection performance.
% %
% On MS COCO, with only prediction mimicking losses applied, our \MyMthd{} boosts the average precision of GFL ResNet-50 with 1$\times$ training schedule from 40.2 to 43.7, outperforming all existing KD methods.
% %
% In addition, our method also works well when distilling detectors with heterogeneous backbones.

Knowledge Distillation (KD) has been validated as an effective model compression technique for learning compact object detectors.
Existing state-of-the-art KD methods for object detection are mostly based on feature imitation.
In this paper, we present a general and effective prediction mimicking distillation scheme, called \MyMthd{}, which delivers the intermediate features of the student's detection head to the teacher's detection head. 
The resulting cross-head predictions are then forced to mimic the teacher's predictions.
This manner relieves the student's head from receiving contradictory supervision signals from the annotations and the teacher's predictions, greatly improving the student's detection performance.
Moreover, as mimicking the teacher's predictions is the target of KD, \MyMthd{} offers more task-oriented information in contrast with feature imitation.
On MS COCO, with only prediction mimicking losses applied, our \MyMthd{} boosts the average precision of GFL ResNet-50 with 1$\times$ training schedule from 40.2 to 43.7, outperforming all existing KD methods.
In addition, our method also works well when distilling detectors with heterogeneous backbones.

\end{abstract}

\section{Introduction}\label{sec:intro}

\begin{figure}[!t]
    \centering
    \setlength{\abovecaptionskip}{3pt}
    \includegraphics[width=\linewidth]{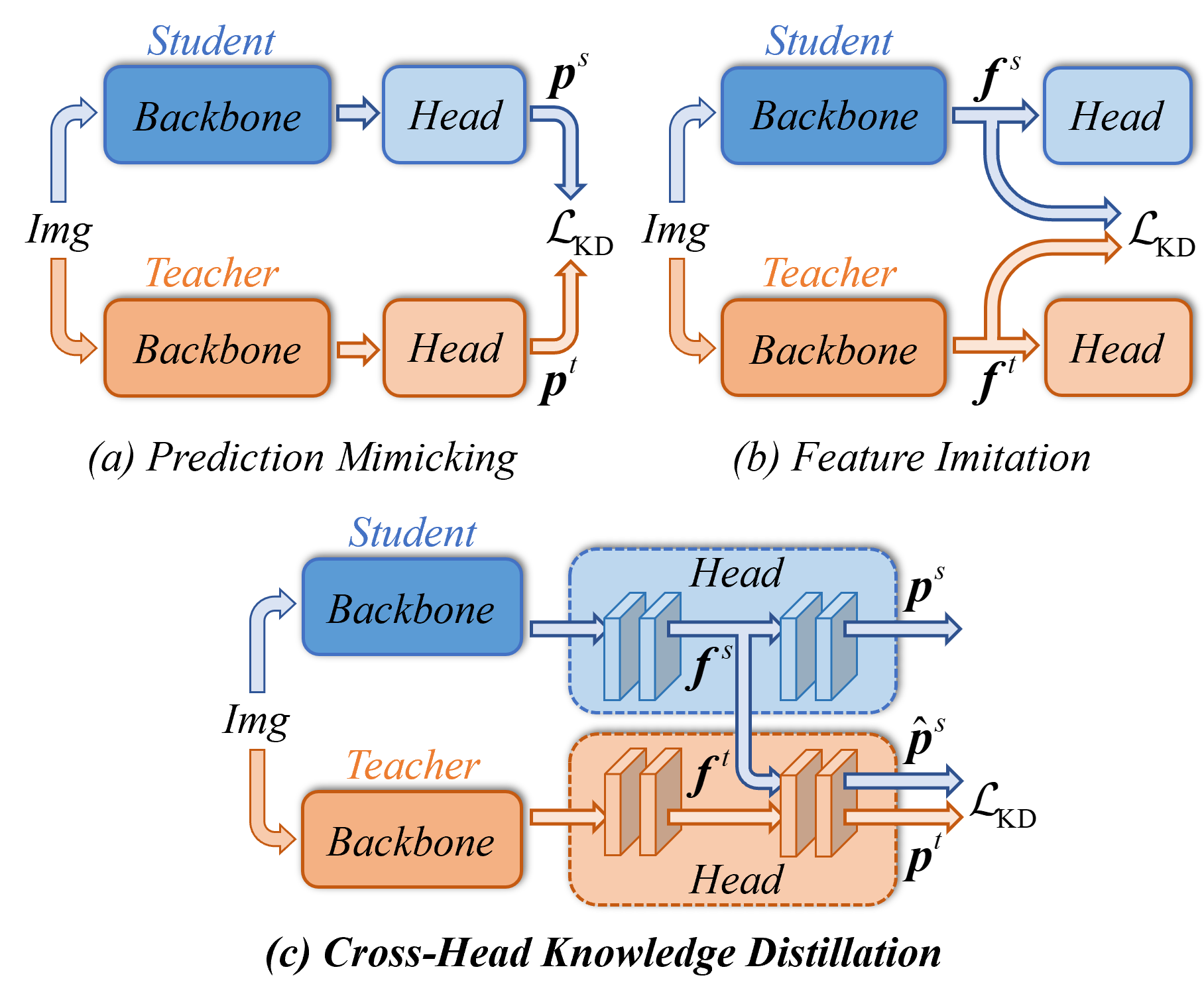}
    \caption{
        Comparisons between conventional KD methods and our \MyMthd.
        Rather than explicitly enforcing the consistency between the intermediate feature maps or the predictions of the teacher-student pair, \MyMthd{} implicitly builds the connection between the heads of the teacher-student pair to improve the distillation efficiency.
    }
    \label{fig:comparison}
    \vspace{-8pt}
\end{figure}

Knowledge Distillation (KD), serving as a model compression technique, has been deeply studied in object detection~\cite{zhixing2021distilling, cao2022pkd, wang2019distilling, dai2021general, li2022knowledge, zheng2022localization,wang2023mvcontrast,song2022dense,lan2022arm3d} and has received excellent performance recently.
According to the distillation position of the detectors, existing KD methods can be roughly classified into two categories: prediction mimicking and feature imitation. 
Prediction mimicking (See \cref{fig:comparison}(a)) was first proposed in~\cite{hinton2015distilling}, which points out that the smooth distribution of the teacher's predictions is more comfortable for the student to learn than the Dirac distribution of the ground truths. 
In other words, prediction mimicking forces the student to resemble the prediction distribution of the teacher. 
Differently, feature imitation (See \cref{fig:comparison}(b)) follows the idea proposed in FitNet~\cite{romero2014fitnets}, which argues that intermediate features contain more information than the predictions from the teacher.
It aims to enforce the feature consistency between the teacher-student pair.

\begin{figure*}
    \centering
    \setlength{\abovecaptionskip}{2pt}
    \includegraphics[width=0.99\textwidth]{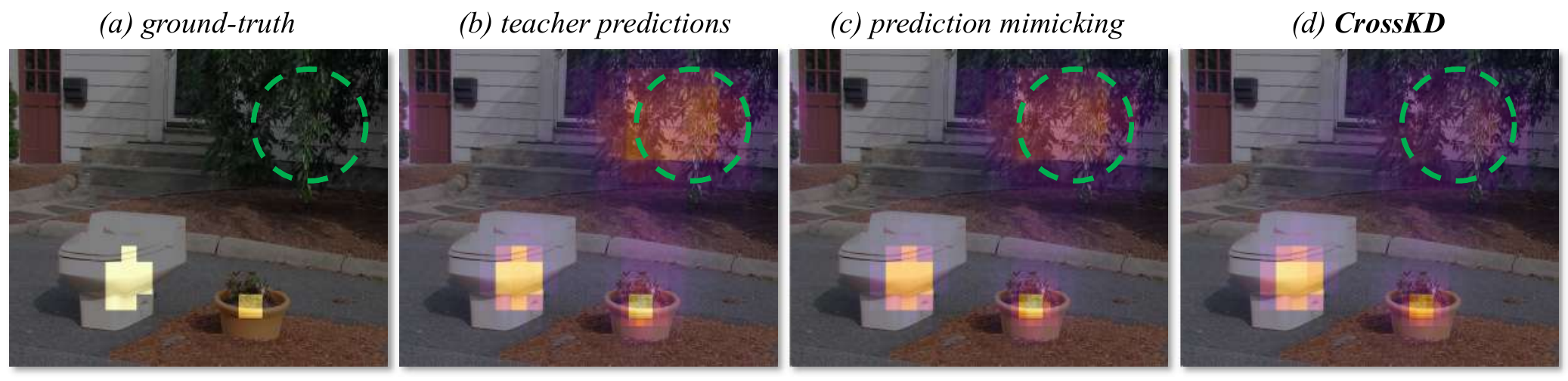}
    \caption{
        Visualizations of the classification predictions from the GFL~\cite{li2020generalized}.
        (a) and (b) are ground truth and distillation targets.
        (c) and (d) are the classification outputs predicted by models training with conventional prediction mimicking and proposed \MyMthd{}.
        In the green circled areas, the distillation targets predicted by the teacher have a large discrepancy with the ground-truth targets assigned to the student.
        prediction mimicking forces the student to mimic the teacher, while \MyMthd{} can smooth the mimicking process.
        }
    \label{fig:problems}
    \vspace{-10pt}
\end{figure*}

Prediction mimicking plays a vital role in distilling object detection models.
However, it has been observed to be inefficient than feature imitation for a long time.
Recently, Zheng et al.~\cite{zheng2022localization} proposed a localization distillation (LD) method that improves prediction mimicking by transferring localization knowledge, which pushes the prediction mimicking to a new level.
%
% Despite still lagging behind the advanced feature imitation methods, e.g., PKD~\cite{cao2022pkd}, LD demonstrates that prediction mimicking has the ability to transfer task-specific knowledge, which benefits the student from the orthogonal aspect to feature imitation.
%
Despite just catching up with the advanced feature imitation methods, e.g., PKD~\cite{cao2022pkd}, LD shows that prediction mimicking has the ability to transfer task-specific knowledge, which benefits the student from the orthogonal aspect to feature imitation.
This motivates us to further explore and improve prediction mimicking.

% prediction mimicking needs to confront the conflict between ground-truth targets and distillation targets, which has been neglected by previous works.
%
Through investigation, we observe that conventional prediction mimicking may suffer from a conflict between the ground-truth targets from the student's assigner and the distillation targets predicted from the teacher.
When training a detector with prediction mimicking, the student's predictions are forced to mimic both the ground-truth targets and the teacher's predictions simultaneously.
However, the distillation targets predicted by the teacher usually have a large discrepancy with the ground-truth targets assigned to the student.
As shown in~\cref{fig:problems}(a) and \cref{fig:problems}(b), the teacher produces class probabilities in the green circled areas, which conflicts with the ground-truth targets assigned to the student.
As a result, the student detector experiences a contradictory learning process during distillation, which seriously interferes with optimization.

% As a result, the student detector experiences a contradictory learning process during distillation, which we argue is the main reason hindering prediction mimicking from achieving higher performance.

To alleviate the above conflict, previous prediction mimicking methods~\cite{zheng2022localization, dai2021general, guo2021distilling} tend to conduct the distillation within regions containing mediate teacher-student discrepancies.
However, we argue that the heavily uncertain regions generally accommodate more information that is beneficial to the student.
In this paper, we present a novel cross-head knowledge distillation pipeline, abbreviated as \emph{\MyMthd{}}.
% In this paper, we present a novel cross-head knowledge distillation pipeline, abbreviated as \emph{\MyMthd{}}, which may alleviates the influence of target conflict problem and enables to mimic the entire predictions from the teacher.
As illustrated in \cref{fig:comparison}(c), We propose to feed the intermediate features from the head of the student to that of the teacher, yielding the cross-head predictions.
Then, the KD operations can be conducted between the new cross-head predictions and the teacher's predictions.

Despite its simplicity, \MyMthd{} offers the following two main advantages.
First, since both the cross-head predictions and the teacher's predictions are produced by sharing part of the teacher's detection head, the cross-head predictions are relatively consistent with the teacher's predictions.
This relieves the discrepancy between the teacher-student pair and enhances the training stability of prediction mimicking.
In addition, as mimicking the teacher’s predictions is the target of KD, \MyMthd{} is theoretically optimal and can offer more task-oriented information compared with feature imitation.
Both advantages enable our \MyMthd{} to efficiently distill knowledge from the teacher's predictions and hence result in even better performance than previous \sArt feature imitation methods.

Without bells and whistles, our method can significantly boost the performance of the student detector, achieving a faster training convergence.
Comprehensive experiments on the COCO \cite{lin2014microsoft} dataset are conducted in this paper to elaborate the effectiveness of \MyMthd.
Specifically, with only prediction mimicking losses applied, \MyMthd{} achieves 43.7 AP on GFL with 1$\times$ training schedule, which is 3.5 AP higher than the baseline, surpassing all previous state-of-the-art object detection KD methods.
Moreover, experiments also indicate our \MyMthd{} is orthogonal to feature imitation methods.
By combining \MyMthd{} with the state-of-the-art feature imitation method, like PKD~\cite{cao2022pkd}, we further achieve 43.9 AP on GFL.
Furthermore, we also show that our method can be used to distill detectors with heterogeneous backbones and performs better than other methods.

\section{Related Work}

\subsection{Object Detection}

Object detection is one of the most fundamental computer vision tasks, which requires recognizing and localizing objects simultaneously.
Modern object detectors can be briefly divided into two categories: one-stage~\cite{lin2017focal, tian2019fcos, zhang2020bridging, li2020generalized, redmon2018yolov3, cao2023domain,chen2023tinydet, chen2023yoloms} detectors and two-stage~\cite{girshick2014rich,girshick2015fast,ren2015faster,lin2017feature,he2017mask, wang2019region, chen2019hybrid, guo2021positive, zhao2024dynamic} detectors.
Among them, one-stage detectors, also known as dense detectors, have emerged as the mainstream trend in detection due to their excellent speed-accuracy trade-off.

Dense object detectors have received great attention since YOLOv1~\cite{redmon2016you}.
Typically, YOLO series detectors~\cite{redmon2016you,redmon2017yolo9000,redmon2018yolov3, bochkovskiy2020yolov4,lyu2022rtmdet,ge2021yolox} attempt to balance the model size and their accuracy to meet the requirement of real-world applications.
Anchor-free detectors~\cite{tian2019fcos, kong2020foveabox, zhu2019feature} attempt to discard the design of anchor boxes to avoid time-consuming box operations and cumbersome hyper-parameter tuning.
Dynamic label assignment methods~\cite{zhang2020bridging, nguyen2022improving, feng2021tood} are proposed to better define the positive and negative samples for model learning.
GFL~\cite{li2020generalized, li2020generalizedv2} introduces Quality Focal Loss (QFL) and a Distribution-Guided Quality Predictor to increase the consistency between the classification score and the localization quality.
It also models the bounding box representation as a probability distribution so that it can capture the localization ambiguity of the box edges.
Recently, attributing to the strong ability of the transformer block to encode expressive features, DETR family~\cite{zhu2020deformable, carion2020end, meng2021conditional, li2022dn, liu2022dabdetr, zhang2022dino, cao2023domain2} has become a new trend in the object detection community.

\subsection{Knowledge Distillation for Object Detection}

Knowledge Distillation (KD) is an effective technique to transfer knowledge from a large-scale teacher model to a small-scale student model.
It has been widely studied in the classification task~\cite{zhao2022decoupled, cho2019efficacy, mirzadeh2020improved, yang2019snapshot, zhang2018deep, romero2014fitnets, heo2019comprehensive, kim2018paraphrasing, park2019relational, yim2017gift, li2021online, li2023curriculum},
but it is still challenging to distill detection models because of the extreme background ratio.
The pioneer work \cite{chen2017learning} proposes the first distillation framework for object detection by simply combining feature imitation and prediction mimicking.
Since then, feature imitation has attracted more and more research attention.
Typically, some works~\cite{li2017mimicking, wang2019distilling, dai2021general, jia2024mssd} focus on selecting effective distillation regions for better feature imitation,
while other works~\cite{guo2021distilling,zhixing2021distilling,li2022knowledge} aim to weight the imitation loss better.
There are also methods~\cite{yao2021g, cao2022pkd, yang2022focal, zhang2021improve} attempting to design new teacher-student consistency functions, aiming to explore more consistency information or release the strict limit of the MSE loss.
%
% Typically, FGD~\cite{yang2022focal} aligns the attention place for the teacher-student pair,
% %
% while PKD~\cite{cao2022pkd} maximizes the Pearson Correlation Coefficient between the feature representations.

As the earliest distillation strategy proposed in~\cite{hinton2015distilling}, prediction mimicking plays a vital role in classification distillation.
Recently, some improved prediction mimicking methods have been proposed to adapt to object detection.
For example, Rank Mimicking~\cite{li2022knowledge} regards the score rank of the teacher as a kind of knowledge and aims to force the student to rank instances as the teacher.
LD~\cite{zheng2022localization} proposes to distill the localization distribution of bounding box~\cite{li2020generalized} to transfer localization knowledge.
%
% Although LD shows the superiority of prediction mimicking to $L_2$ distance-based feature imitation, its accuracy gains are only comparable with the correlation-based feature imitation methods.
%
In this paper, we construct a \MyMthd{} pipeline which separates detection and distillation into different heads to alleviate the target conflict problem of prediction mimicking.
It's worth noting that HEAD~\cite{wang2022head} delivers the student features to an independent assistant head to bridge the gap between heterogeneous teacher-student pairs.
%
% However, the assistant head in HEAD brings additional computational burden and potentially loses information from the teacher.
%
In contrast, we observe that simply delivering the student feature to the teacher is effective enough to achieve SOTA results.
This makes our method quite concise and different from HEAD.
Our method is also related to~\cite{Chaudhuri2019Lit, bai2020few, li2020residual, yang2021knowledge}, but all of them aim to distill classification models and are not tailored for object detection.

\section{Methodology}

\subsection{Analysis of the Target Conflict Problem\label{subsec:problem}}
Target conflict is a common issue confronted in conventional prediction mimicking methods.
In contrast to the classification task, which assigns a specific category to every image, the labels in advanced detectors are usually dynamically assigned and not deterministic.
Typically, detectors depend on a hand-crafted principle, \ie, assigner, to determine the label in each location.
In most cases, detectors cannot reproduce the assigner's labels exactly, which results in a conflict between the teacher-student targets in KD.
Furthermore, the inconsistency of the assigners of the student and teacher in real-world scenarios extends the distance between the ground-truth and distillation targets.

\begin{figure}[!t]
    \centering
    \includegraphics[width=0.96\linewidth]{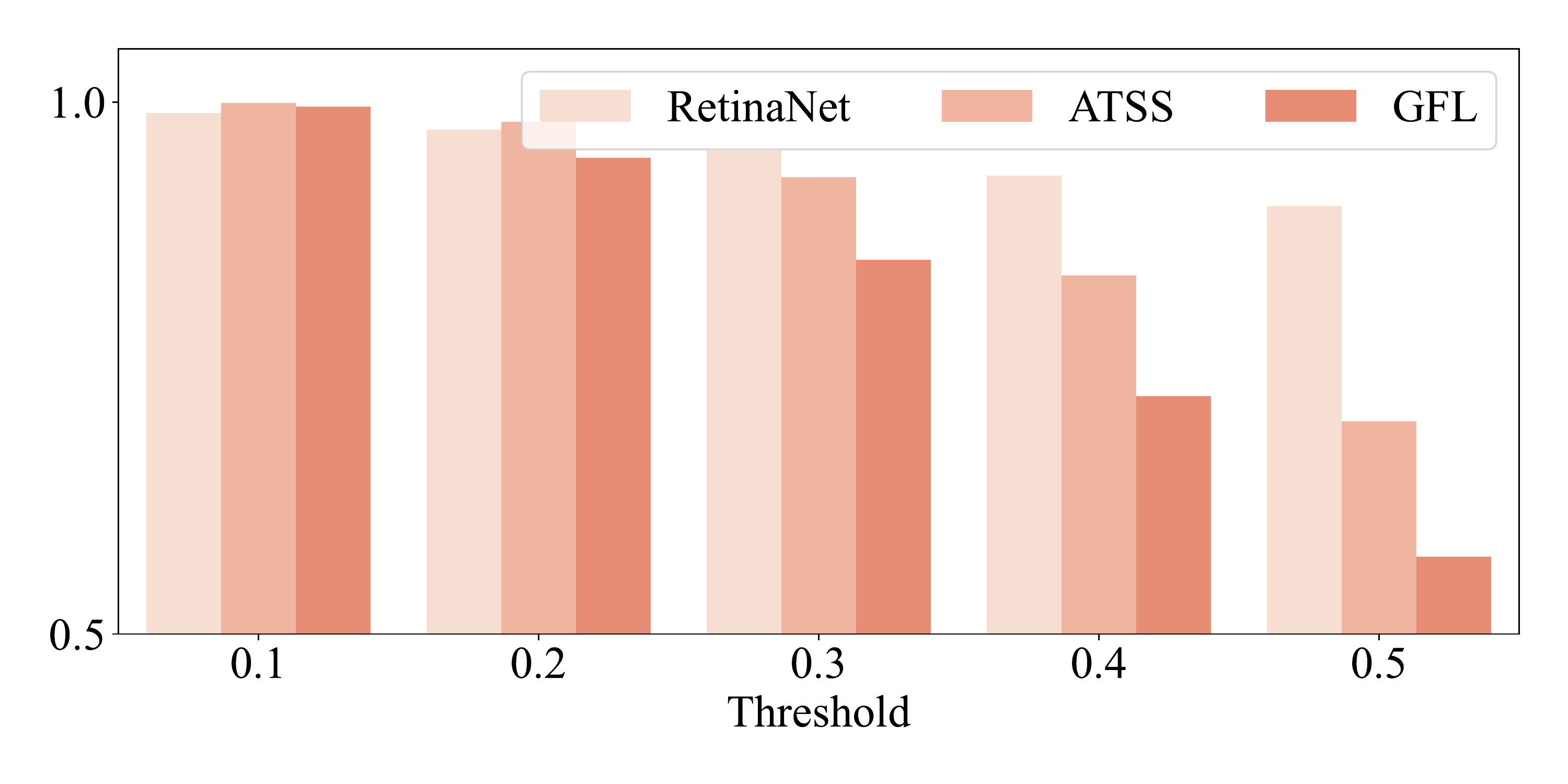}
    \vspace{-15pt}
    \caption{
        Statistics of the target conflict degree between student (GFL-R50) and teacher (GFL-R101, ATSS-R101, RetinaNet-R101).
        X-axis is the teacher-student discrepancy threshold for conflict areas.
        Y-axis represents the ratios of the target conflict areas to the positive areas.
    }
    \label{fig:count}
    \vspace{-10pt}
\end{figure}

To quantitatively measure the degree of target conflict, we statistic the ratios of conflict areas to the positive areas under different teacher-student discrepancy in the COCO \textit{minival} dataset and report the results in \cref{fig:count}.
As we can see, even if both the teacher (ATSS \cite{zhang2020bridging} and GFL~\cite{li2020generalized}) and student (GFL) have the same label assignment strategy, there are still numerous locations that have a discrepancy larger than 0.5 between the ground-truth and distillation targets, respectively.
%
% When we use a teacher with a different label assignment strategy to distill the GFL student, the number of severely-conflict locations increases to 34014，
%
When we use a teacher with a different assigner (RetinaNet) to distill the student (GFL), the conflict areas increases by a large margin.
More experiments in \cref{sec:conflict_exp} also demonstrate that the target conflict problem severely hinders the performance of prediction mimicking.

% \begin{table}[t]
%     \small
%     \centering
%     \setlength{\tabcolsep}{8pt}
%     \setlength{\abovecaptionskip}{2pt}
%     \caption{
%         Numbers of location where the confidence discrepancy between the ground-truth and distillation targets is larger than 0.5.
%     }\label{tab:count}
%     \begin{tabular}{cccc}\toprule[0.8pt]
%         Student & GFL-R50  & GFL-R50   & GFL-R50 \\
%         Teacher & GFL-R101 & ATSS-R101 & RetinaNet-R101 \\ \hline
%         Count  &   3773     &    3986       & 34014 \\ \toprule[0.8pt]
%     \end{tabular}  
%     \vspace{-0.6\baselineskip}
    
% \end{table}

\begin{figure*}[!t]
    \centering
    \includegraphics[width=\textwidth]{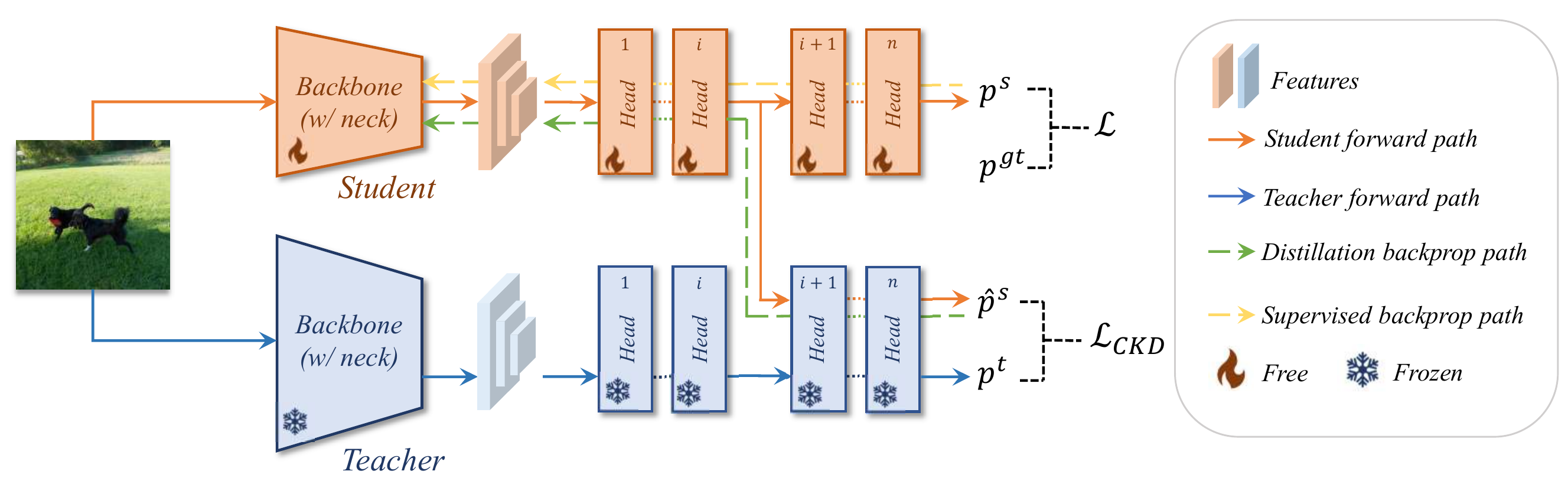}
    \vspace{-20pt}
    \caption{
        Overall framework of the proposed \MyMthd.
        For a given teacher-student pair, \MyMthd~first delivers the intermediate features of the student into
        the teacher layers and generates the cross-head predictions $\hat{\bm{p}}^s$.
        Then, distillation losses are calculated between the original teacher's predictions and the cross-head predictions of the student.
        In back-propagation, the gradients with respect to the detection loss normally pass through the student detection head, while the distillation gradients propagate through the frozen teacher layers.
    }
    \label{fig:structure}
    \vspace{-0.2\baselineskip}
\end{figure*}

Despite the large influence of target conflict, this problem has been neglected for a long time in previous prediction mimicking methods~\cite{hinton2015distilling, li2022knowledge}.
These methods intend to directly minimize the discrepancy between the teacher-student predictions.
Its objective can be described as:
\begin{equation}
    \mathcal{L}_{\text{KD}} = \frac{1}{|\mathcal{S}|}\sum_{r \in \mathcal{R}}{\mathcal{S}(r)\mathcal{D}_{\text{pred}}(\bm{p}^s(r),\bm{p}^t(r))}, \label{eqn:pm}
\end{equation}
where $\bm{p}^s$ and $\bm{p}^t$ are the prediction vectors generated by the detection heads of the student and the teacher, respectively.
$\mathcal{D}_{\text{pred}}(\cdot)$ refers to the loss function calculating the discrepancy between $\bm{p}^s$ and $\bm{p}^t$,
\eg, KL Divergence~\cite{hinton2015distilling} for classification, L1 Loss~\cite{chen2017learning} and LD~\cite{zheng2022localization} for regression.
$\mathcal{S}(\cdot)$ is the region selection principle which produces a weight at each position $r$ in the entire image region $\mathcal{R}$.

It's worth noting that $\mathcal{S}(\cdot)$, to a certain extent, can alleviate the target conflict problem by down-weighting the regions with large teacher-student discrepancies.
However, the heavily uncertain regions usually accommodate more information benefits for the student than undisputed areas.
Ignoring those regions may have a large impact on the effectiveness of prediction mimicking methods.
Consequently, to push the envelope of prediction mimicking, it is necessary to handle the target conflict problem gracefully instead of directly down-weighting.

\subsection{\MyMthdL{} \label{subsec:method}}

As described in \cref{subsec:problem}, we observe that directly mimicking the predictions of the teacher confronts the target conflict problem, which hinders prediction mimicking achieving promising performance.
To alleviate this problem, we present a novel Cross-head Knowledge Distillation (\MyMthd{}) in this section.
The overall framework is illustrated in \cref{fig:structure}.
Like many previous prediction mimicking methods, our \MyMthd{} performs the distillation process on the predictions.
Differently, \MyMthd{} delivers the intermediate features of the student to the teacher's detection head and generates cross-head predictions to conduct distillation.

Given a dense detector, like RetinaNet~\cite{lin2017focal}, each detection head usually consists of a sequence of convolutional layers, represented as $\{C_i\}$.
For simplicity, we suppose there are totally $n$ convolutional layers in each detection head (e.g., 5 in RetinaNet with 4 hidden layers and 1 prediction layer).
We use $\bm{f}_i, i \in \{1, 2, \cdots, n-1\}$ to denote the feature maps produced by $C_i$ and $\bm{f}_0$ the input feature maps of $C_1$.
% %
% In the forward phase, dense detectors successively feed the latent features into convolutional layers $C_i$, described as:
% \begin{equation}
%     \bm{f}_{i} = C_i(\bm{f}_{i-1}), \ i=1,...,n-1.
% \end{equation}
%
The predictions $\bm{p}$ are generated by the last convolutional layer $C_n$.
% , written as:
%
% \begin{equation}
%     \bm{p} = C_\text{n}(\bm{f}_{\text{n}-1}).
% \end{equation}
% %
% To further simplify the expression, we represent the nesting of convolutional layers as $\mathcal{N}_i$, which can be formulated as:
% %
% \begin{equation}
% \begin{aligned}
%     \bm{p} &= C_\text{n}(C_\text{n-1}(\cdots C_{i+1}(\bm{f}_i))) = \mathcal{N}_i(\bm{f}_i).
% \end{aligned}
% \end{equation}
%
% In \MyMthd, We use $\bm{f}^t_i$ and $\bm{f}^s_i$ to denote the feature maps produced by the $i$-th convolutional layer of the teacher and the student, respectively.
% given intermediate features of the teacher and student $\bm{f}^t_i$ $\bm{f}^s_i$, dense detectors successively feed them into convolutional layers and generate the predictions $\bm{p}^t$ $\bm{p}^s$, which can be formulated as:
% %
% \begin{equation}
%     \bm{p}^t = \mathcal{N}^t_i(\bm{f}^t_i), \
%     \bm{p}^s = \mathcal{N}^s_i(\bm{f}^s_i).
% \end{equation}
% %
% Here, $i$ represents the index of the intermediate features in the detection head. $\mathcal{C}^t_i(\cdot)$, $\mathcal{C}^s_i(\cdot)$ indicates the nesting of convolutional layers after the $i$-th features.
%
Thus, for a given teacher-student pair, the predictions of the teacher and the student can be represented as $\bm{p}^t$ and $\bm{p}^s$, respectively.
%
% \begin{equation}
%     \bm{p}^t = \mathcal{N}^t_i(\bm{f}^t_i), \
%     \bm{p}^s = \mathcal{N}^s_i(\bm{f}^s_i).
% \end{equation}
% %
% Here, $\mathcal{N}^t_i$, $\bm{f}^t_i$ and $\mathcal{N}^s_i$, $\bm{f}^s_i$ indicate the convolutional layers and intermediate features of the teacher and the student, repsectively.

Besides the original predictions from the teacher and the student, \MyMthd{} additionally delivers the student's intermediate features $\bm{f}^s_i, i \in \{1, 2, \cdots, n-1\}$ to $C^t_{i+1}$, the $(i+1)$-th convolutional layer of the teacher's detection head, resulting in the cross-head predictions $\hat{\bm{p}}^s$.
%
% a new branch termed cross-head predictions, which can be formulated as:
%
% \begin{gather}
%     \hat{\bm{p}}^s = \mathcal{N}^t_i(\bm{f}^s_i).
% \end{gather}
%
Given $\hat{\bm{p}}^s$, instead of computing the KD loss between $\bm{p}^s$ and $\bm{p}^t$, we propose to use the KD loss
between the cross-head predictions $\hat{\bm{p}}^s$ and the original predictions of the teacher $\bm{p}^t$ as the objective of our \MyMthd{}, which is described as follows:
\begin{equation}
    \begin{aligned}
        \mathcal{L}_{\text{\MyMthd{}}} &= \frac{1}{|\mathcal{S}|}\sum_{r \in \mathcal{R}}{\mathcal{S}(r)\mathcal{D}_\text{pred}(\hat{\bm{p}}^s(r),\bm{p}^t(r))}, \label{eq:kd}
                         % &= \frac{1}{|\mathcal{S}|}\sum_{r \in \mathcal{R}}{\mathcal{S}(r)\mathcal{D}_\text{pred}(\mathcal{N}^t_i(\bm{f}^s_i)(r),\mathcal{N}^t_i(\bm{f}^t_i)(r))} \label{eq:kd},
    \end{aligned} 
\end{equation}
where $\mathcal{S}(\cdot)$ and $|\mathcal{S}|$ are the region selection principle and the normalization factor.
Instead of designing complicated $\mathcal{S}(\cdot)$, we equally conduct distillation between $\hat{\bm{p}}^s$ and $\bm{p}^t$ over the entire prediction map.
Specifically, $\mathcal{S}(\cdot)$ is a constant function with the value of 1 in our \MyMthd{}.
According to the different tasks of each branch (e.g., classification or regression), we perform different types of $\mathcal{D}_{\text{pred}}(\cdot)$ to effectively deliver task-specific knowledge to the student.

By performing \MyMthd{}, the detection loss and the distillation loss are separately applied to different branches. 
As illustrated in \cref{fig:structure}, the gradients of the detection loss pass through the entire head of the student, while the gradients of distillation loss propagate through the frozen teacher layers to the latent features of the student, which heuristically increases the consistency between the teacher and the student.
Compared to directly closing the predictions between the teacher-student pair, \MyMthd{} allows part of the student's detection head to be only relative with detection losses, resulting in a better optimization towards ground-truth targets.
Quantitative analysis is presented in our experiment section.

\subsection{Optimization Objectives \label{subsec:distillationloss}}

The overall loss for training can be formulated as the weighted sum of the detection loss and the distillation loss, written as:
\begin{equation}
    \begin{aligned}
        \mathcal{L} &= \mathcal{L}_\text{cls}(\bm{p}^s_\text{cls},\bm{p}^{gt}_\text{cls}) + \mathcal{L}_\text{reg}(\bm{p}^s_\text{reg},\bm{p}^{gt}_\text{reg}) \\
                    &+ \mathcal{L}^\text{cls}_\text{\MyMthd{}}(\bm{\hat{p}}^s_{\text{cls}},\bm{p}^t_{\text{cls}}) + \mathcal{L}^\text{reg}_\text{\MyMthd{}}(\bm{\hat{p}}^s_\text{reg},\bm{p}^t_\text{reg}),
    \end{aligned}
    \label{eq:loss}
\end{equation}
where $\mathcal{L}_\text{cls}$ and $\mathcal{L}_\text{reg}$ stand for the detection losses which are calculated between the student predictions $\bm{p}^s_\text{cls}$, $\bm{p}^s_\text{reg}$ and their corresponding ground truth targets $\bm{p}^{gt}_\text{cls}$, $\bm{p}^{gt}_\text{reg}$.
The additional \MyMthd{} losses are represented as $\mathcal{L}^\text{cls}_\text{\MyMthd{}}$ and $\mathcal{L}^\text{reg}_\text{\MyMthd{}}$, which are performed between the cross-head predictions $\bm{\hat{p}}^s_\text{cls}$, $\bm{\hat{p}}^s_\text{reg}$ and the teacher's predictions $\bm{p}^t_\text{cls}$, $\bm{p}^t_\text{reg}$.

We apply different distance functions $\mathcal{D}_{\text{pred}}$ to transfer task-specific information in different branches.
In the classification branch, we regard the classification scores predicted by the teacher as the soft labels and directly use Quality Focal Loss (QFL) proposed in GFL~\cite{li2020generalized} to pull close the teacher-student distance.
As for regression, there are mainly two types of regression forms presenting in dense detectors.
The first regression form directly regresses the bounding boxes from the anchor boxes (e.g., RetinaNet~\cite{lin2017focal}, ATSS~\cite{zhang2020bridging}) or points (e.g., FCOS~\cite{tian2019fcos}).
In this case, we directly use GIoU~\cite{rezatofighi2019generalized} as $\mathcal{D}_\text{pred}$.
In the other situation, the regression form predicts a vector to represent the distribution of box location (e.g., GFL~\cite{li2020generalized}), which contains richer information than the Dirac distribution of the bounding box representation.
To efficiently distill the knowledge of location distribution, we employ KL divergence, like LD~\cite{zheng2022localization}, to transfer localization knowledge.
More details about the loss functions are given in the supplementary materials.

\newcommand{\AP}[1]{AP$_{#1}$}
\begin{table}[t!]
  \small
  \centering
  \setlength{\tabcolsep}{8.2pt}
  \setlength{\abovecaptionskip}{2pt}
  \renewcommand{\arraystretch}{1.0}
  \caption{Effectiveness of applying \MyMthd{} at different positions.
    % Effectiveness of the position to apply \MyMthd{}.
    %
    The index $i$ represents the intermediate features used as input in the cross-head branches.
    % $i$ is the index of the intermediate features inputted in the cross-head branches.
    %
    `LD' means the direct application of prediction mimicking on the student's head with LD~\cite{zheng2022localization}.
    % 'PM' means directly performing prediction mimicking on the student's head.
    %
    The teacher-student pair is GFL with ResNet-50 and ResNet-18 backbones.
    We can see that $i=3$ yields the best performance in this experiment.
  }
  \begin{tabular}{ccccccc} \toprule[0.8pt]
    $i$  & AP        &\AP{50}     &\AP{75}     &\AP{S}      &\AP{M}      & \AP{L}    \\ \midrule[0.8pt]
    -    & 35.8      & 53.1       & 38.2       & 18.9       & 38.9       & 47.9      \\ \midrule
    0    & 38.2      & 55.6       & 41.3       & 20.2       & 41.9       & 50.9      \\  
    1    & 38.3      & 55.8       & 41.1       & 20.8       & 42.1       & 49.8      \\
    2    & 38.6      & 56.2       & 41.5       & 20.8       & 42.7       & 50.7      \\ 
    \cc 3& \cc{38.7} & \cc{56.3}  & \cc{41.6}  & \cc{21.1}  & \cc{42.2}  & \cc{51.1} \\ 
    4    & 38.2      & 55.7       & 41.2       & 20.3       & 41.9       & 50.2      \\ \midrule
    LD   & 37.8      & 55.5       & 40.5       & 20.0       & 41.4       & 49.5      \\ \bottomrule[0.8pt]
  \end{tabular} 
  \label{tab:layernum}
\end{table}

\section{Experiments}

\subsection{Implement Details}

We evaluate the proposed method on the large-scale MS COCO~\cite{lin2014microsoft} benchmark as done in most previous works.
To ensure consistency with the standard practice, we use the \textit{trainval135k} set (115\textit{K} images) for training and the \textit{minival} set (5\textit{K} images) for validation. 
% Following the standard practice, all experiments use the \textit{trainval135k} set (115\textit{K} images) for training and \textit{minival} set (5\textit{K} images) as validation.
%
For evaluation, the standard COCO-style measurement, i.e., Average Precision (AP), is used.
We also report mAP with IoU thresholds of 0.5 and 0.75, as well as AP for small, medium, and large objects.
% Standard COCO-style measurement, i.e., Average Precision (AP), is reported as an evaluation metric.
%
Our proposed method, \MyMthd{}, is implemented under the MMDetection~\cite{mmdetection} framework in Python.
% Our \MyMthd{} is implemented based on the MMDetection~\cite{mmdetection} framework in Python.
%
For a fair comparison, all experiments are developed using 8 Nvidia V100 GPUs with a minibatch of two images per GPU. 
% All experiments are developed on 8 Nvidia v100 GPUs with two images on each batch.
%
Unless otherwise stated, all the hyper-parameters follow the default settings of the corresponding student model for both training and testing.
% The training and testing hyper-parameters exactly follow the default protocol of the corresponding student model.

\begin{table}[t!]
    \small
    \centering
    \setlength{\abovecaptionskip}{2pt}
    \setlength{\tabcolsep}{5.8pt}
    \caption{
        Comparisons between feature imitation and \MyMthd{}.
        we choose advanced PKD to represent feature imitation and apply PKD to different positions to compare with \MyMthd{} fairly.
        Here, `neck' means performing PKD on the FPN neck. 'cls' and 'reg' indicate applying PKD to the classification branch and the regression, respectively.
        The teacher-student pair is GFL with ResNet-50 and ResNet-18 backbones.
    }\label{tab:qfeature}
    \begin{tabular}{lcccccc}\toprule[0.8pt]
        % \multirow{2}{*}{}& \multicolumn{3}{c|}{\textbf{RetinaNet}}& \multicolumn{3}{c}{\textbf{GFL}}      \\ 
        Methods        & AP         & \AP{50}    & \AP{75}   & \AP{S}    & \AP{M}    & \AP{L}    \\ \midrule[0.8pt]
        -              & 35.8       & 53.1       & 38.2      & 18.9      & 38.9      & 47.9      \\ \midrule  
        PKD:neck       & 38.0       & 55.0       & 41.2      & 19.6      & 41.5      & 50.2      \\ 
        PKD:cls        & 37.5       & 54.9       & 40.5      & 19.5      & 41.1      & 50.5      \\
        PKD:reg        & 37.2       & 54.0       & 40.2      & 19.0      & 40.9      & 50.0      \\
        PKD:cls+reg    & 37.3       & 54.3       & 40.0      & 19.2      & 41.1      & 49.8      \\ \midrule
        \cc \bMyMthd{} & \cc{38.7}  & \cc{56.3}  & \cc{41.6} & \cc{21.1} & \cc{42.2} & \cc{51.1} \\ \bottomrule[0.8pt]  
    \end{tabular}
    \vspace{.05in}
\end{table}

\subsection{Method Analysis \label{sec:ablation}}

\begin{figure*}[t]
    % \centering
    \setlength{\abovecaptionskip}{0pt}
    \includegraphics[width=\linewidth]{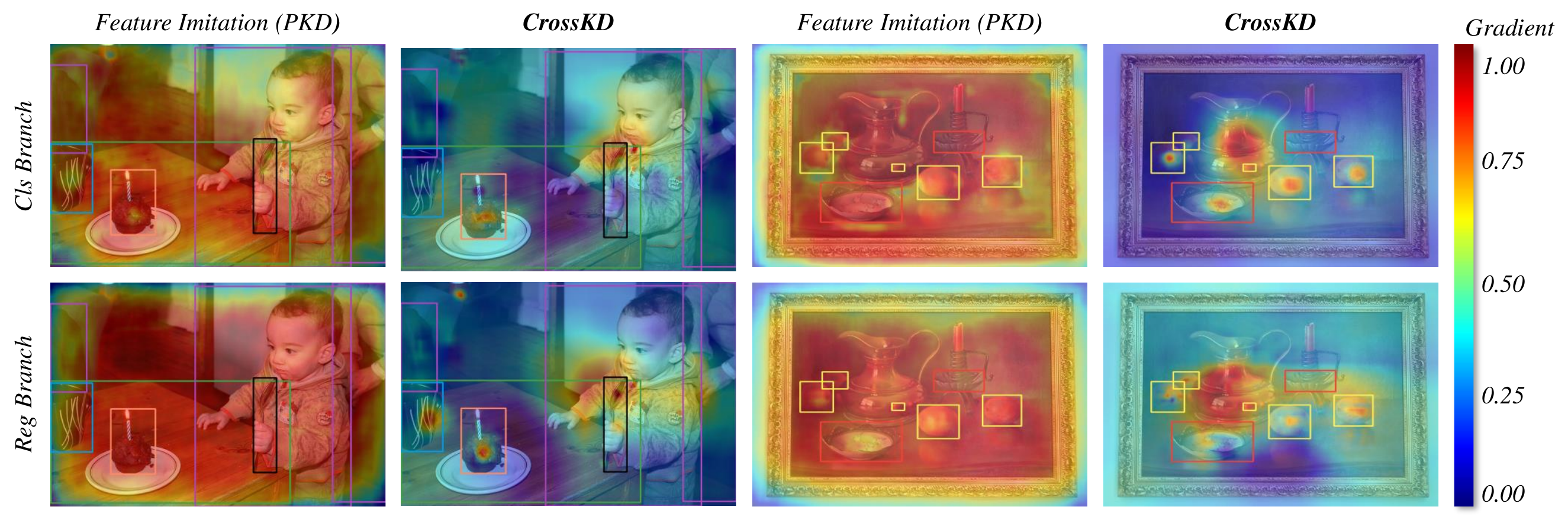}
    % \vspace{-5mm}
    \caption{
        Visualizations of the gradients w.r.t feature imitation and \MyMthd{}.
        The visualization demonstrates that our \MyMthd{} guided by prediction mimicking can effectively focus on the potentially valuable regions.
    }
    % \vspace{-10pt}
    \label{fig:grads}
    \vspace{-10pt}
\end{figure*}

To investigate the effectiveness of our method, we conduct extensive ablation experiments based on GFL~\cite{li2020generalized}.
% We conduct extensive ablation experiments based on  GFL~\cite{li2020generalized} to investigate the effectiveness of our method.
%
If not specified, we use GFL with the ResNet-50 backbone~\cite{he2016deep} as the teacher detector and use the ResNet-18 backbone in the student detector.
% If not specified, we utilize GFL with the ResNet50 backbone as the teacher detector, while the student detector only applies the ResNet18 backbone.
%
The accuracy of the teacher and the student are 40.2 AP and 35.8 AP, respectively.
All experiments follow the default 1$\times$ training schedule (12 epochs). 
% The default 1$\times$ training schedule (12 epochs) is performed in all experiments.

\myPara{Positions to apply \MyMthd{}.}
As described in \cref{subsec:method}, \MyMthd{} delivers the $i$-th intermediate feature of the student to part of the teacher's head.
Here, we conduct distillation on both classification and box regression branches.
When $i=0$, \MyMthd{} directly feeds the student's FPN features into the teacher's head.
In this case, the entire student's head is only supervised by the detection loss, and no distillation loss is involved.
As $i$ gradually increases, more layers of the student's head are jointly affected by the detection loss and the distillation loss.
% As the $i$ gradually increases, more and more layers only affected by the detection loss will also be related to the distillation loss.
%
When $i=n$, our method degrades to the original prediction mimicking, where the distillation loss will be directly performed between the two predictions of the teacher-student pair.
% Finally, the distillation loss will be directly performed between the predictions of teacher-student pair and \MyMthd{} will degenerate into simple prediction mimicking.

In \cref{tab:layernum}, we report the results of performing \MyMthd{} on different intermediate features.
% We list all results when performing \MyMthd{} on different intermediate features in \cref{tab:layernum}.
%
One can see that our \MyMthd{} can improve the distillation performance for all the choices of $i$. %
%This finding implies that the cross-head strategy can effectively enhance the performance of prediction mimicking.
% As the table shows, regardless of the position, \MyMthd{} can boost the results by at least 0.4\% mAP compared with prediction mimicking, which implies that the cross-head strategy can effectively improve the performance of prediction mimicking.
%
Notably, when using the 3-rd intermediate features, \MyMthd{} reaches the best performance of 38.7 AP, which is 0.9 AP higher than the recent state-of-the-art prediction mimicking method LD~\cite{zheng2022localization}.
This suggests that not all layers in the student's head need to be isolated from the influence of the distillation loss.
% It implies that not all layers in student's head need to isolate from the influence of the distillation loss.
%
Therefore, we use $i=3$ as the default setting in all subsequent experiments.
% We use $i=3$ as the default setting in all following experiments.

\begin{table}[t]
    \small
    \centering
    \setlength{\tabcolsep}{6pt}
    \setlength{\abovecaptionskip}{2pt}
    \caption{
        Effectiveness of \MyMthd{} on different branches.
        We separately apply \MyMthd{} on the classification (cls) and regression (reg) branches.
        The teacher-student pair is GFL with ResNet-50 and ResNet-18 backbones.
    }\label{tab:qprediction1}
    \begin{tabular}{cccccccc}
        \toprule[0.8pt]
        \multirow{2}{*}{cls}  & \multirow{2}{*}{reg}  & \multicolumn{3}{c}{LD}         & \multicolumn{3}{c}{\MyMthd{}}     \\  \cmidrule (lr){3-5} \cmidrule (lr){6-8}
                              &                       & AP         & \AP{50}  & \AP{75}  & AP    & \AP{50}     & \AP{75}  \\ \toprule[0.8pt]
        \checkmark            &                       & 37.3       & 55.2     & 40.0     & 37.7      & 55.6       &  40.2     \\ 
                              &\checkmark             & 36.8       & 53.8     & 39.6     & 37.2      & 54.0       &  40.0     \\
        \checkmark            &\checkmark             & 37.8       & 55.4     & 40.5     & \cc{38.7} & \cc{56.3}  & \cc{41.6}     \\ \bottomrule[0.8pt]
    \end{tabular}
    \vspace{-10pt}      

\end{table}

\myPara{\MyMthd{} v.s. Feature Imitation.\label{subsec:vsfeat}}
We compare \MyMthd{} with the advanced feature imitation method PKD~\cite{cao2022pkd}.
For a fair comparison, we perform PKD on the same positions as our \MyMthd{}, including FPN features and the third layer of detection heads.
The results are reported in \cref{tab:qfeature}.
% For a fair comparison, we separately perform PKD on the FPN features and the features which are delivered to teacher's head in \MyMthd{}.
%
It can be seen that PKD can achieve 38.0 AP when it is applied between the FPN features of the teacher-student pair.
On the detection head, PKD even shows a performance drop.
%only receives 34.9 AP when applied on the classification branch, 37.2 AP when applied on the regression branch, and 37.3 AP when applied on both branches.
%
In contrast, our \MyMthd{} achieves 38.7 AP, which is 0.7 AP higher than PKD applied on the FPN features.
%
%From the results, \MyMthd{} is more effective than feature imitation, particularly on the detection head.

To further investigate the advantage of \MyMthd{}, we visualize the gradients on the latent features of the detection head, as shown in \cref{fig:grads}.
%  \red{To further investigate the advantage of \MyMthd{}, we calculate the gradient among channels on the latent features of the detection head and visualize it as heatmap, as shown in \cref{fig:grads}.} 
%
As illustrated, the gradients generated by PKD have a large and wide impact on the entire feature maps, which is inefficient and not targeted.
On the contrary, the gradients generated by \MyMthd{} can focus on potential semantic areas with objects of interest.
%
%This may be mainly because the gradients with respect to \MyMthd{} are calculated by prediction mimicking loss, which has a clear physical meaning and can offer task-specific knowledge.

% In addition, we further conduct a visualization for the pre-pixel $l_1$ distance between student-teacher model pairs in \cref{fig:distance}. 
% %
% As shown in \cref{fig:distance}, the prediction result of the student exhibits a variation from the teacher's when using feature imitation.
% %
% On the contrary, our \MyMthd~reduces the distance between student and teacher, leading to an efficient distillation.

\begin{table}[t]
    \small
    \centering
    \setlength{\tabcolsep}{5pt}
    \setlength{\abovecaptionskip}{2pt}
    \caption{
        Collective effect of \MyMthd{} and prediction mimicking.
        The teacher-student pair is GFL with ResNet-50 and ResNet-18 backbones.
    }\label{tab:qprediction2}
    \begin{tabular}{cccccccc}
        \toprule[0.8pt]
            \MyMthd{}  & LD           & AP         & \AP{50}  & \AP{75}  & \AP{S}  & \AP{M}  & \AP{L}  \\ \midrule[0.8pt]
            -          & -            & 35.8       & 53.1       & 38.2       & 18.9      & 38.9      & 47.9      \\ \midrule
            \checkmark &              & \cc{38.7}  & \cc{56.3}  & \cc{41.6}  & \cc{21.1} & \cc{42.2} & \cc{51.1} \\
                       & \checkmark   & 37.8       & 55.5       & 40.5       & 20.0      & 41.4      & 49.5      \\ 
            \checkmark & \checkmark   & 38.1       & 55.6       & 40.9       & 20.4      & 41.6      & 51.1 \\
        \bottomrule[0.8pt]          
    \end{tabular}  
    \vspace{-10pt}
\end{table}

\begin{figure*}[t]
    \centering
    \includegraphics[width=0.9\textwidth]{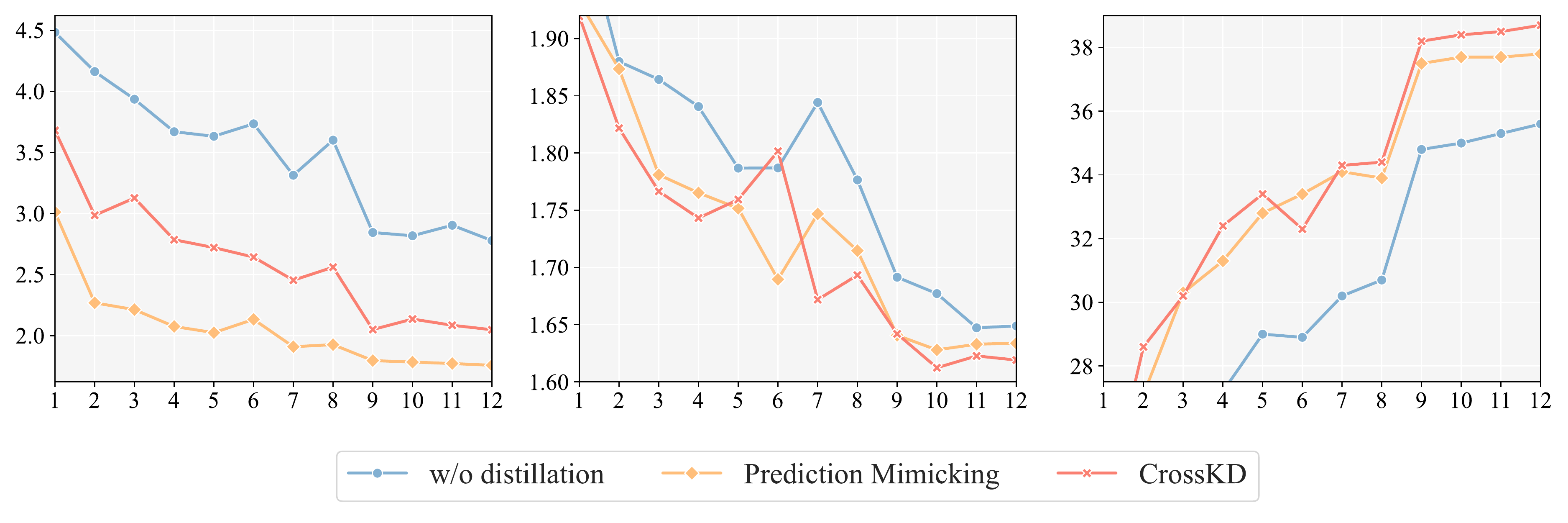}
    % \begin{overpic}
    \put(-435, 150){\footnotesize (a) Avg. $L_1$ distance between $\bm{p}^s$ and $\bm{p}^t$}
    \put(-290, 150){\footnotesize (b) Avg. $L_1$ distance between $\bm{p}^s$ and $\bm{p}^{gt}$}
    \put(-80, 150){\footnotesize (c) AP}
    % \end{overpic}
    \setlength{\abovecaptionskip}{3pt}
    \caption{
        Visualization for the variation of statistics during training. 
        (a) Curves of average $L_1$ distance between student predictions $\bm{p}^s$ and teacher's $\bm{p}^t$.
        (b) Curves of average $L_1$ distance between student predictions $\bm{p}^s$ and positive ground truth targets $\bm{p}^{gt}$.
        (c) Curves of Average Precision (AP).
        All curves are evaluated on the COCO \textit{minival} set.
        X-axis refers to the epoch number.
        Y-axis in (a) and (b) indicate the average $L_1$ distance, while in (c) means the value of AP.
    }
    \label{fig:error_bar}
    \vspace{-10pt}
\end{figure*}

\myPara{\MyMthd{} v.s. Prediction Mimicking.}
%
%In this section, we aim to showcase the advantages of \MyMthd{} by comparing it with prediction mimicking.
% In this subsection, we further compare \MyMthd{} with prediction mimicking to demonstrate the advantages of the cross-head strategy.
%
We begin by separately performing prediction mimicking and \MyMthd{} on the classification and box regression branches.
% We first separately perform directly prediction mimicking and \MyMthd{} on the classification branch and regression branch.
%
The results are reported in \cref{tab:qprediction1}.
One can see that replacing prediction mimicking with \MyMthd{} leads to a stable performance gain regardless of classification or regression branches.
% As the results show, replacing prediction mimicking with \MyMthd{} can bring a stable accuracy gain regardless of classification or regression branches.
%
Specifically, prediction mimicking produces 37.3 AP and 36.8 AP on the classification and regression branches, respectively,
while \MyMthd{} yields 37.7 AP and 37.2 AP, representing a consistent improvement over the corresponding results of prediction mimicking.
% 
% Meanwhile, \MyMthd{} achieves 37.7\% mAP and 37.2\% mAP, which are 0.4\% mAP higher than corresponding results of prediction mimicking.
%
If KD is performed on the two branches, our method can still outperform prediction mimicking by +0.9 AP.
% When performing KD on both branches, predition mimicking can only achieves 37.8\% mAP while \MyMthd{} achieves 38.7\% mAP.
%
Moreover, we further evaluate the collective effect of prediction mimicking and \MyMthd{}, as shown in \cref{tab:qprediction2}.
% Furthermore, we simultaneously apply prediction mimicking and \MyMthd{} to evaluate the collective effect of prediction mimicking and \MyMthd{}.
%
Intriguingly, we observe that using both prediction mimicking and \MyMthd{} together yields a final result of 38.1 AP, which is even lower than the result of using \MyMthd{} alone. 
% As evident from the \cref{tab:qprediction} (b), using both prediction mimicking and \MyMthd{} yields a final result of 38.1 AP, which is lower than single \MyMthd{}.
% 
We believe that this is because the prediction mimicking introduces the target conflict problem again, which makes the student model struggle to learn.
% We consider it is because the prediction mimicking introduces the target conflict problem again, which hinders student's accuracy.

In addition, we visualize the statistical variation during training to conduct further analysis on \MyMthd{} and prediction mimicking.
We first calculate the $L_1$ distances between the student's predictions $\bm{p}^s$ and the teacher's predictions $\bm{p}^t$, as well as the ground-truth targets $\bm{p}^{gt}$ at each epoch.
As plotted in ~\cref{fig:error_bar}(a), the distance $L_1(\bm{p}^s,\bm{p}^t)$ can be reduced significantly by our \MyMthd{},
while it is reasonable for the prediction mimicking to achieve the lowest distance as the distillation is directly imposed on $\bm{p}^s$. 
%is relatively mild when closing the teacher-student distance, as the distance between $p^s$ and $p^t$ is less than the case without distillation but higher than the direct prediction mimicking throughout the training process.
%
However, as the optimization target conflict exists, the prediction mimicking involves a contradictory optimization process, thereby generally yielding a larger distance $L_1(\bm{p}^s,\bm{p}^{gt})$ than our \MyMthd{}, as shown in ~\cref{fig:error_bar}(b).
In \cref{fig:error_bar}(c), our method shows a faster training process and achieves the best performance of 37.8 AP.

\def\FitNets{FitNets*~\cite{romero2014fitnets}        & 40.7 (0.5$\uparrow$)}
\def\InsideG{Inside GT Box*                           & 40.7 (0.5$\uparrow$)}
\def\Defeat{Defeat*~\cite{guo2021distilling}          & 40.8 (0.6$\uparrow$)}
\def\MainRe{Main Region*~\cite{zheng2022localization} & 41.1 (0.9$\uparrow$)}
\def\FineGr{Fine-Grained*~\cite{Wang_2019_CVPR}       & 41.1 (0.9$\uparrow$)}
\def\FGD{FGD~\cite{yang2022focal}                     & 41.3 (1.1$\uparrow$)}
\def\GID{GID*~\cite{dai2021general}                   & 41.5 (1.3$\uparrow$)}
\def\LD{LD~\cite{zheng2022localization}               & 43.0 (2.8$\uparrow$)}
\def\LDd{LD$\dagger$~\cite{zheng2022localization}     & xx.x (x.x$\uparrow$)}
\def\SKD{SKD~\cite{de2022structural}                  & 42.3 (2.1$\uparrow$)}
\def\PKD{PKD*~\cite{cao2022pkd}                       & 43.3 (3.1$\uparrow$)}
\def\myRslt{\textbf{\MyMthd}                          & 43.7 (3.5$\uparrow$)}
\def\myRsltP{\textbf{\MyMthd+PKD} &\fg{{43.9}} (\textbf{\fg{3.7$\uparrow$}})}

\begin{table}[t]
    \small
    \centering
    \setlength{\tabcolsep}{2.3pt}
    \renewcommand\arraystretch{1.1}
    \setlength{\abovecaptionskip}{3pt}
    \caption{
        Comparison with \sArt detection KD methods on COCO.
        * denotes results are referenced from LD~\cite{zheng2022localization} and PKD~\cite{cao2022pkd}.
        All results are evaluated on the COCO \textit{minival} set.
    }\label{tab:sotakd}
    \begin{tabular}{lcccccc} \toprule[0.8pt]
        Method & AP &\AP{50} &\AP{75} &\AP{S} &\AP{M} &\AP{L} \\ \midrule[0.8pt]
        GFL-R101 (T) & 44.9 & 63.1 & 49.0   & 28.0  & 49.1  & 57.2 \\
        GFL-R50 (S)  & 40.2 & 58.4 & 43.3   & 23.3  & 44.0  & 52.2 \\ \midrule
        \FitNets & 58.6 & 44.0 & 23.7 & 44.4 & 53.2 \\
        \InsideG & 58.6 & 44.2 & 23.1 & 44.5 & 53.5 \\
        \Defeat  & 58.6 & 44.2 & 24.3 & 44.6 & 53.7 \\
        \MainRe  & 58.7 & 44.4 & 24.1 & 44.6 & 53.6 \\
        \FineGr  & 58.8 & 44.8 & 23.3 & 45.4 & 53.1 \\
        \FGD     & 58.8 & 44.8 & 24.5 & 45.6 & 53.0 \\
        \GID     & 59.6 & 45.2 & 24.3 & 45.7 & 53.6 \\
        % \LD      & xx.x & xx.x & xx.x & xx.x & xx.x \\
        \SKD     & 60.2 & 45.9 & 24.4 & 46.7 & 55.6 \\  
        \LD      & 61.6 & 46.6 & 25.5 & 47.0 & 55.8 \\
        \PKD     & 61.3 & 46.9 & 25.2 & 47.9 & 56.2 \\  \midrule \rowcolor[gray]{.95}
        \myRslt  & \fg{62.1} & 47.4 & \fg{26.9} & 48.0 & 56.2 \\  \rowcolor[gray]{.95}
        \myRsltP & 62.0 & \fg{47.7} & 26.4 & \fg{48.5} & \fg{57.0} \\ \bottomrule[0.8pt]       
    \end{tabular}  
    \vspace{-.05in} 
\end{table}

\subsection{Comparison with SOTA KD Methods}

In this section, we evaluate various \sArt object detection KD methods on the GFL~\cite{li2020generalized} framework and fairly compare them with our proposed \MyMthd{}. 
% In this section, we adopt varieties of \sArt object detection KD methods on GFL~\cite{li2020generalized} and fairly compare with our \MyMthd{}.
%
We use ResNet-101 as the backbone for the teacher detector, which is trained with a 2$\times$ schedule and multi-scale augmentation.
For the student detector, we adopt the ResNet-50 backbone.
We train the student with the 1$\times$ schedule.
% The teacher detector is based on the ResNet101-FPN backbone and trained with the 2$\times$ schedule and multi-scale augmentation, as well as the student only adopts the ResNet50-FPN backbone and is trained with the 1$\times$ schedule.
%
The pre-trained checkpoint of the teacher is directly borrowed from the MMDetection\cite{mmdetection} model zoo.

\begin{table}[t]
    \small
    \centering
    \setlength{\tabcolsep}{2.5pt}
    \setlength{\abovecaptionskip}{3pt}
    \caption{
        \MyMthd{} for detectors with homogeneous backbones.
        Teacher detectors use ResNet-101 as the backbone, while the students use ResNet-50 as the backbone.
        All results are evaluated on the COCO \textit{minival} set.
    }\label{tab:homogeneous}
    \begin{tabular}{cccccccc} \toprule[0.8pt]
        Student  & Methods    & AP     & \AP{50}  & \AP{75} & \AP{S} & \AP{M} & \AP{L} \\ \midrule[0.8pt]
        \multirow{3}{*}{RetinaNet~\cite{lin2017focal}}  & R101       & 38.9   & 58.0     & 41.5    &21.0    & 32.8   & 52.4   \\  
        & R50        & 37.4   & 56.7     & 39.6    &20.0    & 40.7   & 49.7   \\  %\cmidrule{2-8}
        & \cc \bMyMthd{} & \cc 39.7   & \cc 58.9     & \cc 42.5    & \cc 22.4    & \cc 43.6   & \cc 52.8   \\  \midrule[0.8pt]
        \multirow{3}{*}{FCOS~\cite{tian2019fcos}}       & R101       & 40.8   & 60.0     & 44.0    &24.2    & 44.3   & 52.4   \\ 
        & R50        & 38.5   & 57.7     & 41.0    &21.9    & 42.8   & 48.6   \\  %\cmidrule{2-8}
        & \cc \bMyMthd{} & \cc 41.3   & \cc 60.6     & \cc 44.2    & \cc 25.1    & \cc 45.5   & \cc 52.4   \\  \midrule[0.8pt]            
        \multirow{3}{*}{ATSS~\cite{zhang2020bridging}}  & R101       & 41.5   & 59.9     & 45.2    &24.2    & 45.9   & 53.3   \\ 
        & R50        & 39.4   & 57.6     & 42.8    &23.6    & 42.9   & 50.3   \\  %\cmidrule{2-8}
        & \cc \bMyMthd{} & \cc 41.8   & \cc 60.1     & \cc 45.4    & \cc 24.9    & \cc 45.9   & \cc 54.2   \\ \bottomrule[0.8pt]     
    \end{tabular}  
    \vspace{-.05in}

\end{table}

We report all results in \cref{tab:sotakd}.
As we can see, at the same condition, \MyMthd{} can achieve 43.7 AP without bells and whistles, which improves the accuracy of the student by 3.5 AP, outperforming all other \sArt methods.
Notably, \MyMthd{} surpasses the advanced feature imitation method PKD by 0.4 AP and surpasses the advanced prediction mimicking method LD by 0.7 AP, demonstrating the effectiveness of \MyMthd{}. 
In addition, we also observe that \MyMthd{} is also orthogonal to the feature imitation methods.
With the help of PKD, \MyMthd{} achieves the highest results of 43.9 AP, achieving an improvement of 3.7 AP compared to the baseline.

\subsection{\MyMthd{} on Different Detectors} 

Besides performing \MyMthd{} on GFL, we select three commonly used detectors, i.e., RetinaNet\cite{lin2017focal}, FCOS~\cite{tian2019fcos}, and ATSS~\cite{zhang2020bridging}, to investigate the effectiveness of \MyMthd{}.
We strictly follow the student settings for training and reference the teacher and student results from the MMDetection model zoo. 
The results are presented in \cref{tab:homogeneous}.
% \cref{tab:homogeneous} presents the results of adopting our \MyMthd{} on different detectors.
% Here, the teacher and student results are referenced from MMDetection model zoo, and the training schedule strictly follows the student setting.
%
As shown in \cref{tab:homogeneous}, \MyMthd{} significantly boosts the performance of all three types of detectors.
% As shown in \cref{tab:homogeneous}, \MyMthd{} significantly boosts the performance of all kinds of detectors.
%
Specifically, RetinaNet, FCOS, and ATSS with our \MyMthd{} achieve 39.7 AP, 41.3 AP, and 41.8 AP, respectively, which are 2.3 AP, 2.8 AP, and 2.4 AP higher than their corresponding baselines.
All results after distillation even outperform the original teachers, indicating that \MyMthd{} can work well on different dense detectors.
% All results after distillation are even better than the original teachers, which demonstrates \MyMthd{} works well on different dense detectors.

\begin{table}[t!]
    \small
    \centering
    \setlength{\tabcolsep}{3pt}
    \setlength{\abovecaptionskip}{3pt}
    \caption{
        \MyMthd{} for teacher-student pairs with different label assigners.
        All results are evaluated on the COCO \textit{minival} set.
    }\label{tab:conflict}
    \begin{tabular}{ccccccc}
        \toprule[0.8pt]
            Methods                                          & AP            & \AP{50}     & \AP{75}    & \AP{S}     & \AP{M}       & \AP{L}            \\ \midrule[0.8pt]
            GFL-R50 (S)                                      & 40.2          & 58.4        & 43.3       & 23.3       & 44.0         & 52.2              \\ 
            ATSS\cite{zhang2020bridging}-R101 (T)            & 41.5          & 59.9        & 45.2       & 24.2       & 45.9         & 53.3              \\ \midrule
            KD                                               & 39.7          & 57.9        & 42.8       & 21.8       & 44.2         & 51.5              \\  
            \bMyMthd{}                                       & 42.1          & 60.5        & 45.7       & 24.5       & 46.3         & 54.5              \\ \midrule[0.8pt] 
            GFL-R50 (S)                                      & 40.2          & 58.4        & 43.3       & 23.3       & 44.0         & 52.2              \\ 
            Retinanet\cite{lin2017focal}-R101 (T)            & 38.9          & 58.0        & 41.5       & 21.0       & 32.8         & 52.4              \\ \midrule
            KD                                               & 30.3          & 49.2        & 31.2       & 20.0       & 38.1         & 34.4              \\  
            \bMyMthd{}                                       & 41.2          & 59.4        & 44.8       & 24.0       & 45.1         & 53.5              \\ \midrule[0.8pt]
            % GFL-R50 $\spadesuit$ (S)                         & 45.1          & 65.4        & 48.2       & 30.0       & 49.7         & 54.7      \\ 
            % GFL\cite{li2020generalized}-R101 (T)             & 44.9          & 63.1        & 49.0       & 28.0       & 49.1         & 57.2      \\ \midrule
            % KD                                               & 48.4          & 68.7        & 52.1       & 33.2       & 54.6         & 59.1      \\  
            % \bMyMthd{}                                       & 48.8          & 69.1        & 52.5       & 33.4       & 54.1         & 59.5      \\ \midrule[0.8pt]
        \end{tabular}
    \vspace{-.05in}
\end{table}

\subsection{Distillation under Severe Target Conflict}\label{sec:conflict_exp}
In this subsection, we perform prediction mimicking and our \MyMthd{} between detectors with different assigners to explore the effectiveness of \MyMthd{} against the target conflict problem.
As shown in \cref{tab:conflict}, the target conflict problem has a large impact on the optimization of the student, leading to an inferior performance.
Specifically, prediction mimicking reduces the AP to 30.3 with the teacher as RetinaNet which has a different assigner with GFL.
Furthermore, even if the ATSS has the same assigner as GFL, the student's AP is only distilled to 39.7,  falling below the performance without KD. 
In contrast, \MyMthd{} can still significantly improve the student's accuracy even if existing a large discrepancies between the ground-truth and distillation targets.
\MyMthd{} boosts the accuracy of GFL-R50 to 42.1 (+1.9 AP) when applying ATSS as the teacher.
Even guided by a weak teacher ReitnaNet, \MyMthd{} still improves the performance of GFL-R50 to 41.2 AP, 1.0 AP higher than the baseline.
This demonstrates how robust our \MyMthd{} is when confronting severe target conflict.

\subsection{Distillation between Heterogeneous Backbones}

In this subsection, we explore the ability of our \MyMthd{} for distilling the heterogeneous students.
% Our \MyMthd{} can also be flexibly adopted on detectors with heterogeneous backbones.
%
We perform knowledge distillation on RetinaNet~\cite{lin2017focal} with different backbone networks and compare our method with the recent state-of-the-art method PKD~\cite{cao2022pkd}.
% To explore the capacity of distilling knowledge from heterogeneous backbones, we perform \MyMthd{} between RetinaNet~\cite{lin2017focal} with different series backbones and compare our method with PKD~\cite{cao2022pkd}. 
%
Specifically, we choose two typical heterogeneous backbones, i.e., the transformer backbone Swin-T~\cite{Liu_2021_ICCV} and the lightweight backbone MobileNetv2~\cite{Sandler_2018_CVPR}.
All the detectors are trained for 12 epochs with a single-scale strategy.
The results are presented in \cref{tab:heterogeneous}.
%
% As shown in \cref{tab:heterogeneous}, the improvements obtained by our \MyMthd{} are substantial.
%
We can see when distilling knowledge from Swin-T, \MyMthd{} reaches 38.0 AP (+1.5 AP), outperforming PKD by 0.8 AP.
% improving the student's performance by 1.5 AP.
%
\MyMthd{} also improves the results of RetinaNet with the MoblieNetv2 backbone to 34.1 AP, which is 3.2 AP higher than the baseline and surpasses PKD by 0.9 AP.
%
% These experiments indicate that our \MyMthd{} can perform well for distilling knowledge between detectors with heterogeneous backbones as well.

\begin{table}[t!]
    \small
    \centering
    \setlength{\tabcolsep}{3.65pt}
    \setlength{\abovecaptionskip}{3pt}
    \caption{
        \MyMthd{} for other detector pairs with  Heterogeneous Backbones.
        For convenience, only the backbone lists below, where `SwinT' refers to RetinaNet with a tiny version of Swin Transformer~\cite{Liu_2021_ICCV}.
        All results are evaluated on the COCO \textit{minival} set.
    }\label{tab:heterogeneous}
    \begin{tabular}{ccccccc}
        \toprule[0.8pt]
            Methods                                 & AP            & \AP{50}     & \AP{75}    & \AP{S}     & \AP{M}       & \AP{L}     \\ \midrule[0.8pt]
            SwinT (T)~\cite{Liu_2021_ICCV}          & 37.3          & 57.5        & 39.9       & 22.7       & 41.0         & 49.6       \\ 
            ResNet-50 (S)                           & 36.5          & 55.4        & 39.1       & 20.4       & 40.3         & 48.1       \\ \midrule
            PKD                                     & 37.2          & 56.7        & 39.5       & 21.2       & 41.2         & 49.7       \\ \rowcolor[gray]{.95}            
            \bMyMthd{}                              & 38.0          & 58.1        & 40.5       & 23.1       & 41.8         & 49.7       \\ \midrule[0.8pt] 
            ResNet-50 (T)                           & 36.5          & 55.4        & 39.1       & 20.4       & 40.3         & 48.1       \\ 
            MobileNetv2 (S)~\cite{Sandler_2018_CVPR}& 30.9          & 48.7        & 32.5       & 16.3       & 33.5         & 41.9       \\ \midrule
            PKD                                     & 33.2          & 51.3        & 35.0       & 16.5       & 36.6         & 46.5       \\ \rowcolor[gray]{.95}              
            \bMyMthd{}                              & 34.1          & 52.7        & 36.5       & 18.8       & 37.1         & 45.4       \\ \bottomrule[0.8pt]
    \end{tabular}
    \vspace{-.05in}
\end{table}

\section{Conclusions and Discussions}

In this paper, we introduce \MyMthd{}, a novel KD method designed to enhance the performance of dense object detectors.
\MyMthd{} transfers the intermediate features from the student's head to that of the teacher to produce the cross-head predictions for distillation, an efficient way to alleviate the conflict between the supervised and distillation targets.
% , whilst reliefs the tough constraint of prediction mimicking.
% The distillation loss of \MyMthd{} does not affect on gradient back propagation process for the end part of the student's head, which helps to alleviate the conflict between supervised and distillation targets.
% In addition, the tough constraint of prediction mimicking is relieved by the consistency between the predictions of cross-head and teacher, due to the sharing part of the teacher's head.
Our results have shown that \MyMthd{} can improve the distillation efficiency and achieve \sArt performance.
%
% We hope our work can provide the community with new insights.
%
% In the future, we will further extend our method to two-stage RCNN series, end-to-end detectors (DETR series), and other relevant fields, \eg 3D object detection. 
%
In the future, we will further extend our method to other relevant fields, \eg 3D object detection. 
%
% Besides, our \MyMthd{} can be further explored to the two-stage detectors (RCNN series), sparse detectors (DETR series), and other relevant fields, \eg 3D object detection, rotated object detection, and instance segmentation. 

\myPara{Acknowledgments.}
This research was supported by NSFC (NO. 62225604, NO. 62276145), the Fundamental Research Funds for the Central Universities (Nankai University, 070-63223049), CAST through
Young Elite Scientist Sponsorship Program (No. YESS20210377). Computations were supported by the Supercomputing Center of Nankai University (NKSC).

% \section*{Acknowledgments}

%%%%%%%%% REFERENCES
{\small
\bibliographystyle{ieee_fullname}
\bibliography{egbib}
}

\clearpage
\section*{Appendix}
\section{Details of Distillation Losses}

According to the task of detection heads, \ie, classification, and regression, we apply different distance functions $\mathcal{D}_{\text{pred}}$ to transfer task-specific information in different branches.
In this section, we introduce the details of distance functions $\mathcal{D}_{\text{pred}}$ applied in \MyMthd{}.

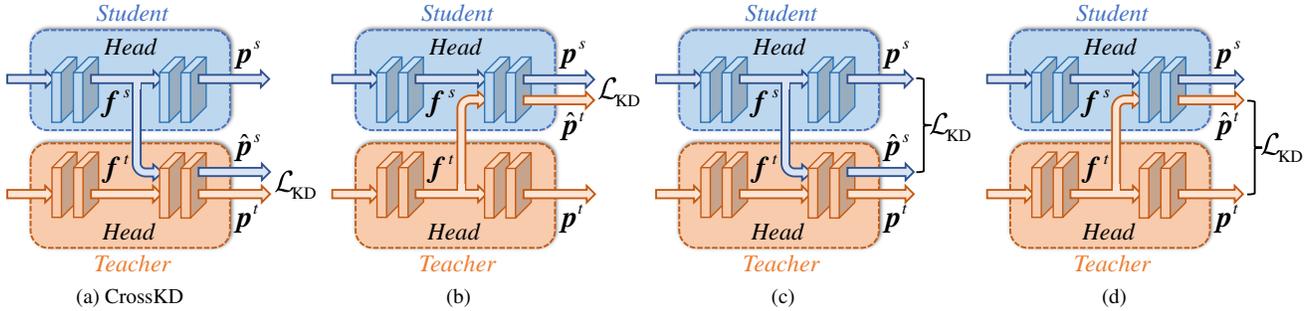
\begin{figure*}[t]
    \centering
    \includegraphics[width=\textwidth]{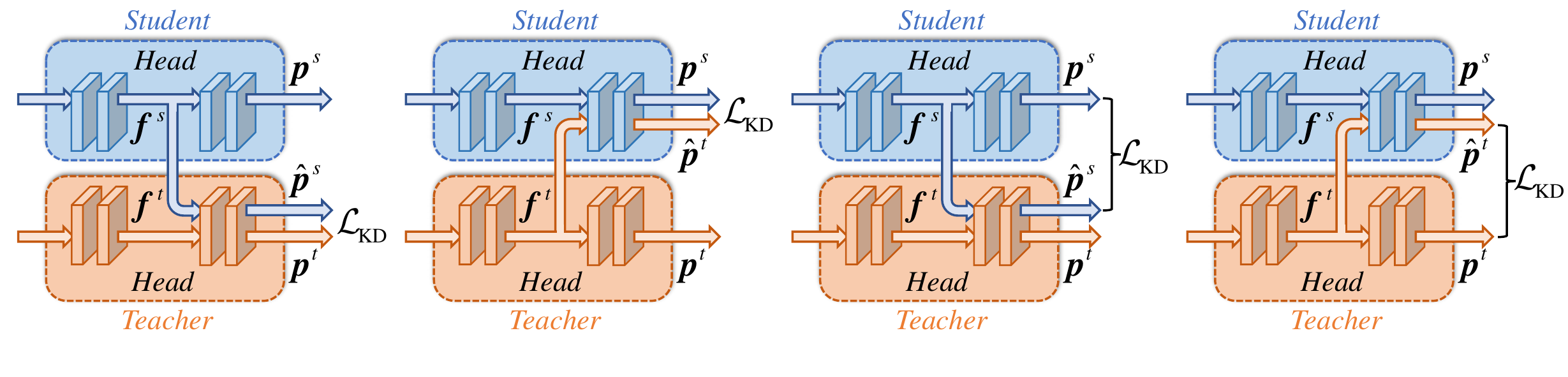}
    % \begin{overpic}
    \put(-465, 3){\footnotesize (a) \MyMthd{}}
    \put(-325, 3){\footnotesize (b)}
    \put(-202, 3){\footnotesize (c)}
    \put(-77, 3){\footnotesize (d)}
    % \end{overpic}
    \setlength{\abovecaptionskip}{3pt}
    \caption{
        Different cross-head strategies.
        (a) is the original strategy used in CrossKD.
        (b) delivers the intermediate features of the teacher to the student head and conducts KD between the cross-head predictions of the teacher and the student's predictions.
        (c) does the same cross-head strategy as (a) but performs KD between the student's original predictions and cross-head predictions.
        (d) does the same cross-head strategy as (b) but performs KD between the teacher's original predictions and the cross-head predictions.
    }
    \label{fig:others}
    % \vspace{0.2\baselineskip}
\end{figure*}

\myPara{Regression Branch.}
There are mainly two types of regression branches that existed in dense detectors.
The first regression branch directly regresses the bounding boxes from the anchor boxes (\eg, RetinaNet~\cite{lin2017focal}, ATSS~\cite{zhang2020bridging}) or points (\eg, FCOS~\cite{tian2019fcos}).
In this case, we directly use GIoU~\cite{rezatofighi2019generalized} as $\mathcal{D}_\text{pred}$, which is defined as:
\begin{equation}
\begin{aligned}
    \mathcal{D}_\text{pred}(\mathcal{B}, \mathcal{B}^{\prime}) = \frac{\vert \mathcal{B} \cap \mathcal{B}^{\prime} \vert}{\vert \mathcal{B} \cup \mathcal{B}^{\prime} \vert} - \frac{\vert \mathcal{C} \setminus (\mathcal{B} \cup \mathcal{B}^{\prime})) \vert}{\vert \mathcal{C} \vert}, \label{eq:Dreg1}
\end{aligned}
\end{equation}
where $\mathcal{B}$ and $\mathcal{B}^{\prime}$ represent the predicted and ground-truth bounding boxes and $\mathcal{C}$ is the smallest enclosing convex object for $\mathcal{B}$ and $\mathcal{B}^{\prime}$. 

In the other situation, the regression branch predicts a vector to represent the distribution of box location (\eg, GFL~\cite{li2020generalized}), which contains richer information than the Dirac distribution of the bounding box representation.
To efficiently distill the knowledge of location distribution, we employ the same $\mathcal{D}_\text{pred}$ like LD~\cite{zheng2022localization}, which is defined as:
\begin{equation}
    \mathcal{D}_{\text{pred}}(\bm{p}, \bm{p}^{\prime}) =  \text{KL}(s(\bm{p}/\tau), s(\bm{p}^{\prime}/\tau)), \label{eq:Dreg2}
\end{equation}
where $\text{KL}$ means KL divergence, $s(\cdot)$ indicates the Softmax function, and $\tau$ is a factor to smooth the distribution.

\myPara{Classification Branch.} 
Distillation in the classification branch severely suffers from the imbalance of the foreground and background instances problem.
To avoid training crash, previous prediction mimicking methods usually design complicated region selection principle to choose effective areas.
In contrast, without selecting effective regions, we regard the classification scores predicted by the teacher as the soft labels and directly use Quality Focal Loss (QFL) proposed in GFL~\cite{li2020generalized} to pull close the teacher-student distance.
We define $\mathcal{D}_\text{pred}$ in the classification branch as:
\begin{equation}
    \mathcal{D}_{\text{pred}}(\bm{p}, \bm{p}^{\prime}) = (|\sigma(\bm{p})-\sigma(\bm{p}^{\prime})|)^\gamma \cdot \text{BCE}(\sigma(\bm{p}), \sigma(\bm{p}^{\prime})), \label{eq:Dcls}
\end{equation}
where $\sigma$ denotes the sigmoid function and $\text{BCE}$ indicates binary cross entropy. 
$(|\sigma(\bm{p})-\sigma(\bm{p}^{\prime})|)^\gamma$ serves as a modulating factor added to the cross entropy function, with a tunable focusing parameter $\gamma \geq 0$.
Here, $\gamma$ is set as 1 in all experiments, which we find is the optimum.

We also compare the performance of QFL with the widely used BCE loss.
As shown in \cref{tab:loss}, The BCE loss can receive 36.3 and 36.2 AP when separately applied on the positive and negative regions.
When we perform distillation on both positive and negative regions, BCE loss can only achieve 36.9 AP, far below 38.7 AP of QFL, which demonstrates the effectiveness of the current distillation losses.

\begin{table}[t]
    \small
    \centering
    \setlength{\tabcolsep}{4.5pt}
    \begin{tabular}{cccccccc}
        \toprule[0.8pt]
        Loss          & Region & AP    & \AP{50}  & \AP{75}  & \AP{S} & \AP{M}   & \AP{L}   \\ \toprule[0.8pt]
        -             & -      & 35.8  & 53.1     & 38.2     & 18.9   & 38.9     & 47.9      \\ 
        BCE           & P      & 36.3  & 53.8     & 39.1     & 19.1   & 39.6     & 48.3      \\ 
        BCE           & N      & 36.2  & 53.5     & 38.9     & 19.3   & 40.0     & 48.2      \\
        BCE           & P+N    & 36.9  & 54.3     & 39.5     & 20.0   & 40.7     & 48.4      \\
        QFL           & P+N    & 38.7  & 56.3     & 41.6     & 21.1   & 42.2     & 51.5      \\ \bottomrule[0.8pt]
    \end{tabular}
    \vspace{.05in}      
    \caption{
        Effectiveness of different distillation losses in classification branch.
        `BCE' and `QFL' means the binary cross entropy loss and quality focal loss, respectively.
        `P' and `N' refer to the positive and negative regions.
        The teacher-student pair is GFL with ResNet-50 and ResNet-18 backbones.
        }
    \label{tab:loss}
\end{table}

\section{The Generalization Ability of \MyMthd{}}
\MyMthd{} is adaptable for any detector distillation since the target conflict is a common problem of object detection distillation due to imperfect teacher predictions.
To demonstrate the generalization, we apply \MyMthd{} on detectors with various types of backbones and structures.

\newcommand{\rNet}[1]{\multirow{2}{*}{ResNet-#1}}
\begin{table}[t]
  \small
  \centering
  \setlength{\tabcolsep}{3.3pt}
  \begin{tabular}{cccccccc} \toprule[0.8pt]
    Student&\bMyMthd& AP &\AP{50}&\AP{75}&\AP{S}&\AP{M}&\AP{L} \\ \midrule[0.8pt]
    \rNet{18} &  & 35.8 & 53.1 & 38.2 & 18.9 & 38.9 & 47.9 \\ 
    & \checkmark & \cc{39.2} & \cc{57.0} & \cc{42.2} & \cc{22.7} & \cc{43.0} & \cc{51.3} \\ \midrule
    \rNet{34} &  & 38.9 & 56.6 & 42.2 & 21.5 & 42.8 & 51.4 \\ 
    & \checkmark & \cc{42.4} & \cc{60.4} & \cc{45.8} & \cc{24.4} & \cc{46.8} & \cc{55.6} \\ \midrule
    \rNet{50} &  & 40.2 & 58.4 & 43.3 & 23.3 & 44.0 & 52.2 \\ 
    & \checkmark & \cc{43.7} & \cc{62.1} & \cc{47.4} & \cc{26.9} & \cc{48.0} & \cc{56.2} \\ 
    \bottomrule[0.8pt] 
  \end{tabular}  
  % \vspace{.05in}
  \caption{Quantitative results of \MyMthd{} for lightweight detectors.
    Standard 1$\times$ schedule is applied in all experiments.
    The teacher detector is GFL with ResNet-101 backbones.
  }\label{tab:lightweight}
\end{table}

The results of our \MyMthd{} on a series of lightweight students distilled with GFL with ResNet-18, ResNet-34, and ResNet-50 backbones are presented in \cref{tab:lightweight}. 
We apply ResNet-101 as the backbone for the teacher detector.
As shown in \cref{tab:lightweight}, our method can effectively enhance the performance of all given lightweight detectors.
Specifically, \MyMthd{} achieves stable improvements for the students with ResNet-18, ResNet-34, and ResNet-50 backbones, which reach 39.2 AP, 42.4 AP, and 43.7 AP.

Furthermore, we adapt \MyMthd{} to typical Faster R-CNN (two-stage) and Deformable DETR (DETR-like) detectors and report their performance in \cref{tab:generalization}.
In Faster R-CNN, we deliver the student region features to the R-CNN head of the teacher to generate cross-head predictions to accept the teacher's supervision.
In Deformable DETR, the cross-head predictions are created by passing the encoder features of the student into each stage of the teacher decoder.
As shown in \cref{tab:generalization}, without finely tuned hyper-parameters, \MyMthd{} boosts the accuracy of ResNet-18 based Faster R-CNN and Deformable DETR to 35.5 (2.0 $\uparrow$) and 45.8 (1.7$\uparrow$) AP, which demonstrates the generalization ability of \MyMthd{}.

\begin{table}[htp!]
    \setlength{\tabcolsep}{4pt}
    \small
    \begin{tabular}{lcccc}
        \toprule[0.6pt]
            Method                   & Schedule & \AP{}                           & \AP{50}  & \AP{75}   \\ \toprule[0.8pt]
            Faster R-CNN R18 (S)     & 12e      & 33.5                            & 53.7     & 35.9      \\ 
            Faster R-CNN R50 (T)     & 12e      & 37.4                            & 58.1     & 40.4      \\ \midrule 
            \MyMthd{}                & 12e      & \textbf{35.5} (2.0$\uparrow$)   & 55.8     & 38.0      \\ \midrule
            % Deformable DETR R18 (S) & 12e      & 38.5                            & 56.6     & 41.2      \\ 
            Deform. DETR R18 (S)     & 50e      & 44.1                            & 62.8     & 47.9      \\ 
            Deform. DETR R50 (T)     & 50e      & 47.0                            & 66.1     & 50.9      \\ \midrule
          % \MyMthd{}-student query   & 12e      & 39.0                            & 55.9     & 41.9      \\
            % \MyMthd{}               & 12e      & 40.9 (2.4$\uparrow$) & 58.1     & 44.1      \\
            \MyMthd{}               & 50e       & \textbf{45.8} (1.7$\uparrow$)   & 63.8     & 49.9      \\
        \bottomrule[0.8pt]
    \end{tabular}
    \caption{CrossKD for Faster R-CNN and Deformable DETR.}
    \label{tab:generalization}
\end{table}

\section{More Ablations}

\begin{table}[h]
    \small
    \centering
    \setlength{\tabcolsep}{6.5pt}
    \begin{tabular}{ccccccc}
        \toprule[0.8pt]
        Strategy    & AP    & \AP{50}  & \AP{75}  & \AP{S}  & \AP{M}   & \AP{L}   \\ \toprule[0.8pt]
        -           & 35.8  & 53.1     & 38.2     & 18.9    & 38.9     & 47.9     \\ 
        (a)         & 38.7  & 56.3     & 41.6     & 21.1    & 42.2     & 51.5     \\ 
        (b)         & 35.4  & 52.5     & 37.8     & 18.6    & 38.4     & 47.1     \\
        (c)         & 34.5  & 51.9     & 36.7     & 17.8    & 37.6     & 45.1     \\ 
        (d)         & 32.5  & 48.8     & 35.0     & 16.6    & 35.0     & 42.8     \\
        \bottomrule[0.8pt]
    \end{tabular}
    \vspace{.05in}      
    \caption{
        Comparisons of different cross-head strategies.
        The strategies (a), (b), (c), (d) have shown in~\cref{fig:others}, where (a) is the current strategy used in \MyMthd{}.
        The teacher-student pair is GFL with ResNet-50 and ResNet-18 backbones.
        }
    \label{tab:cross-head}
\end{table}

In this section, we experiment different cross-head strategies to demonstrate the effectiveness of our \MyMthd{}, which are illustrated in~\cref{fig:others}. 
% We also apply different cross-head strategies as shown in~\cref{fig:others}.
%
As presented in~\cref{tab:cross-head}, strategy (b), which differently reuses the student's detection head, achieved only 35.4 AP, significantly lower than the 38.7 AP obtained by \MyMthd{}.
% As presented in the~\cref{tab:cross-head}, strategy (b), which differently reuses the student's detection head, only achieves 35.3 AP, significantly lower than the 38.7 AP of \MyMthd{}.
%
We hypothesize that this difference in performance may be attributed to the suboptimal optimization of the student's blocks in this approach.
% We argue that it may be blamed on no optimization of the student's blocks.
%
\cref{fig:others}(c) and \cref{fig:others}(d) minimize the distances between the original predictions and the cross-head predictions.
However, these strategies have limited impact on the student's backbones, resulting in 34.5 AP and 32.5 AP for \cref{fig:others}(c) and \cref{fig:others}(d), respectively.
% 
% Due to the limited influence to the backbones of the student, \cref{fig:others}(c) and \cref{fig:others}(d) can only receive 34.5 and 32.5 AP, respectively.

Moreover, \cref{fig:others}(b), (c), and (d) all perform distillation losses and detection losses at the student's detection heads, so the target conflict problem still exists.
In contrast, \MyMthd{} separates the distillation losses onto the teacher's branch and hence avoids the target conflict problem.
As a result, 
% In this way, 
\MyMthd{} receives the highest AP of 38.7 among all cross-head strategies.
% By decoupling the distillation and detection losses, \MyMthd{} receives the highest 38.7 AP among all cross-head strategies, which demonstrates its effectiveness.

\section{Relation to Previous Works}

In this section, we describe the differences of our method and some related works which are originally designed for the classification task~\cite{Chaudhuri2019Lit, bai2020few, li2020residual, yang2021knowledge, liufunction}.
Here, we compare \MyMthd{} with these works from the aspects of motivation and structure to emphasize the differences.

\myPara{Motivation.}
Previous works all concentrate on the classification task.
For instance, Bai \etal~\cite{bai2020few} aims to alleviate overfitting in few-shot task.
Li \etal~\cite{li2020residual} focuses on using a residual network to help a non-residual network overcome gradient vanishing.
Some works~\cite{liufunction, yang2021knowledge} target on the general KD scenario in classification.
These methods all attempts to solve specific problems in classification and are not specially designed for distilling object detectors.
% These methods are not specially designed for distilling detectors.
%
% Therefore, they can not achieve a state-of-the-art performance in the object detection KD scenario.

In contrast, \MyMthd{}, which is specially designed for the object detection task, focuses on the target conflict problem in object detection.
To our knowledge, this is the first work to discuss the target conflict problem in distilling object detectors.
As presented in Sec. 1 of the main paper, the teacher detector usually predicts inaccurate results, which conflict with the ground-truth targets. 
The traditional KD methods supervise the student detector with those two controversial labels at the same place, resulting in low distillation efficiency.
To alleviate this problem, we propose to deliver the intermediate features of the student to the part of the teacher's detection head and generate new cross-head predictions to accept the distillation losses.
%
% In this way, \MyMthd{} relieves the target conflict problem and hence achieves a state-of-the-art performance with only prediction mimicking applied.

However, without the detection-specific design, those methods can not achieve a promising performance.

\myPara{Structure.}
Previous works tend to design a complicated manner to utilize the teacher-student latent features.
Typically, Li \etal~\cite{li2020residual} forwards every stage features of the student into the teacher's blocks.
Liu \etal~\cite{liufunction} alternately delivers the intermediate features from the student to the teacher or from the teacher to the student.
%
% These strategies not only increase the computational complexity but also are hard to extend to other tasks.
These strategies significantly increase the computational complexity in training phase.

Instead of applying a complicated design, \MyMthd{} is relatively simple, which only passes the student's latent features through part of the teacher's detection head.
Despite its simplicity, extensive experiments demonstrate its effectiveness in object detection KD.

\section{Result Visualization}

\begin{figure*}[t]
    \centering
    \includegraphics[width=\textwidth]{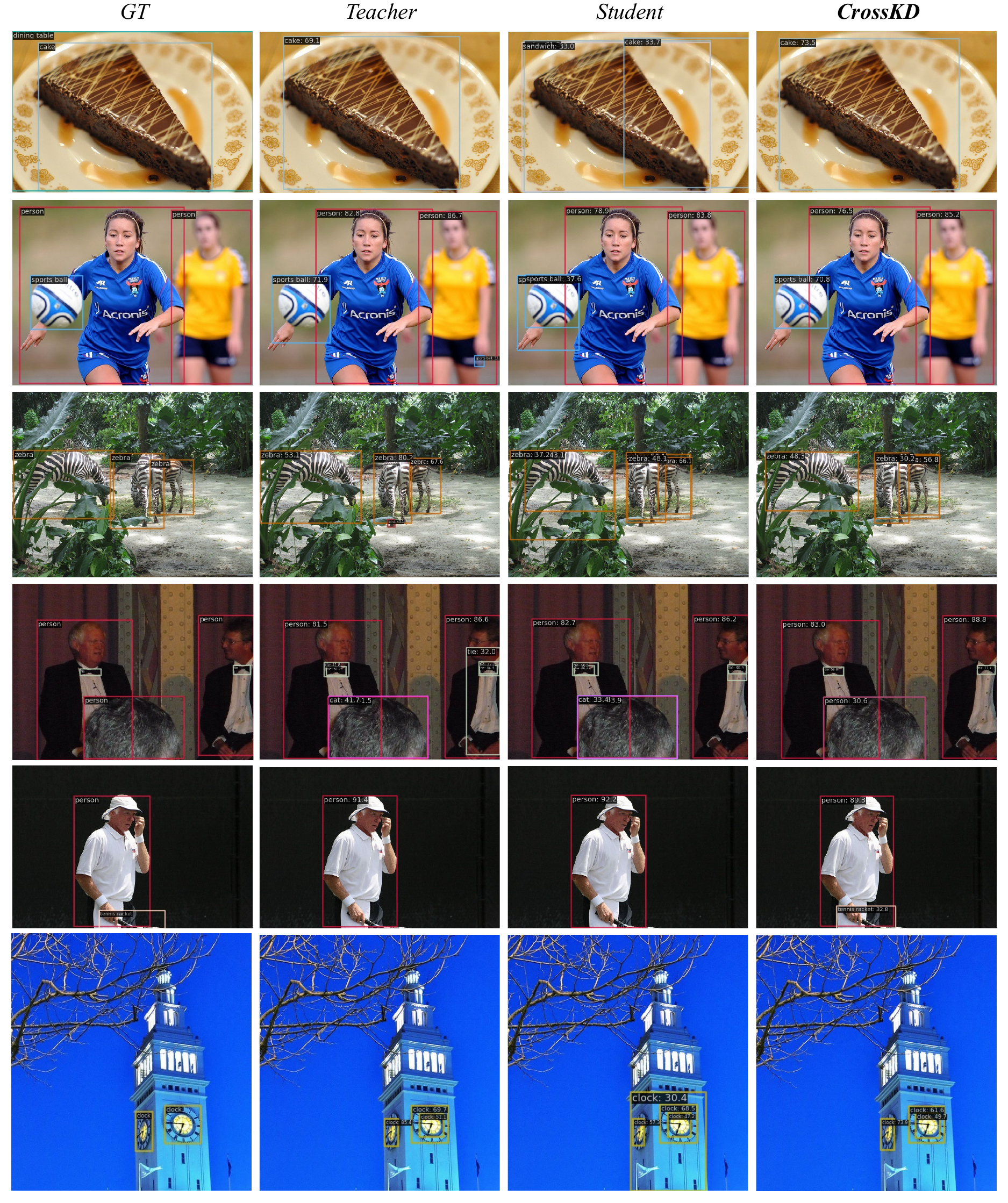}
    \setlength{\abovecaptionskip}{3pt}
    \caption{
        Visualization of detection results of \MyMthd{}.
        The teacher is GFL-R50 with 40.2 AP and student is GFL-R18 with 35.8 AP.
    }
    \label{fig:result}
\end{figure*}

We visualize the detection results of the teacher, the student, and our \MyMthd{} in \cref{fig:result}.
As the visualization shows, \MyMthd{} usually receives even better results than the teacher detector, which demonstrates that \MyMthd{} can relieve the influence of the teacher's inaccurate predictions and achieve a better optimization towards ground-truths.

\end{document}